\documentclass{svjlour3}
\usepackage[utf8]{inputenc}
\usepackage{caption}
\usepackage{subcaption}
\usepackage{graphicx}
\usepackage{float}
\usepackage{amsmath, fullpage}
\usepackage{tabto}
\usepackage{xcolor}
\usepackage{adjustbox}
\usepackage{amssymb} % for \checkmark
\usepackage{siunitx}
\usepackage{booktabs}
\usepackage[numbers,sort&compress]{natbib}
\usepackage[colorlinks=true,allcolors=blue]{hyperref}
\RequirePackage{fix-cm}

\definecolor{DiffRed}{RGB}{255,200,200}
\definecolor{DiffGreen}{RGB}{200,255,200}

\newcommand{\hlr}[1]{\begingroup\fboxsep=0.6pt\colorbox{DiffRed}{#1}\endgroup}
\newcommand{\hlg}[1]{\begingroup\fboxsep=0.6pt\colorbox{DiffGreen}{#1}\endgroup}

\usepackage{titlesec}

\titlespacing*{\section}{0pt}{1.3ex}{0.6ex}
\titlespacing*{\subsection}{0pt}{1.3ex}{0.5ex}
\titlespacing*{\subsubsection}{0pt}{0.8ex}{0.4ex}
\titleformat{\section}{\bfseries\large}{\thesection}{1em}{}
\titleformat{\subsection}{\bfseries\normalsize}{\thesubsection}{1em}{}
\titleformat{\subsubsection}{\bfseries\normalsize}{\thesubsubsection}{1em}{}

\title{HUGO-CS: A Hybrid-Labeled, Uncertainty-Aware, General-Purpose, Observational Dataset for Cold Spray}

\author{Stephen Price$^{1,*}$ \and
        Kyle Miller$^{4}$ \and
        Marco Musto$^{4}$ \and
        Kenneth Kroenlein$^{4}$ \and
        James Saal$^{4}$ \and
        Kyle Tsaknopoulos$^{3}$ \and
        Elke A. Rundensteiner$^{2}$ \and
        Danielle L. Cote$^{3}$}

\institute{
$^{1}$ Computer Science Department, Worcester Polytechnic Institute, Worcester, MA, USA \\
$^{2}$ Data Science Department, Worcester Polytechnic Institute, Worcester, MA, USA \\
$^{3}$ Materials and Manufacturing Department, Worcester Polytechnic Institute, Worcester, MA, USA \\
$^{4}$ Citrine Informatics, Redwood City, CA, USA \\
$^{*}$ Corresponding author: \email{sprice@wpi.edu}
}

\date{Received: \today / Accepted: date}
\begin{document}
\maketitle
\begin{abstract}
Cold spraying is an increasingly common approach for repairing and manufacturing components due to its solid-state manufacturing capabilities. However, process optimization remains difficult due to many interdependent parameters and the lack of large-scale, machine-readable data to support modeling. While the  scientific literature contains many relevant experiments, results are inconsistently reported (often in tables and figures) and use non-uniform units, limiting utilization at scale. To address these limitations, this work presents HUGO-CS, a literature-derived dataset of 4,383 cold-spray experiments with 144 features from 1,124 sources, exceeding the previous largest dataset (137 samples) by 30x. With completely manual extraction requiring an average of 91 minutes per document, this work designs and leverages a \textbf{H}ybrid-labeled, \textbf{U}ncertainty-aware, \textbf{G}eneral-purpose, \textbf{O}bservational extraction framework, called HUGO, to support this extraction. HUGO combines automated LLM-based labeling with targeted manual label refinement to handle this experimental result extraction process from scientific literature. To balance labeling efficiency with extraction accuracy, HUGO introduces a Hierarchical Risk Mitigation (HRM) to route LLM outputs with a high risk of potential errors for manual review, while retaining low-risk records as auto-labeled. Lastly, HUGO post-processing consolidates categorical descriptors, maps reported feedstock chemistries into structured continuous compositions, and normalizes units across sources. Of the 4,383 reported experiments, 1,765 are hand-labeled, providing a high-quality labeled subset for benchmarking, error analysis, and higher-fidelity data points. All code to replicate this work, along with the complete HUGO-CS dataset, are released under a CC-BY license at \href{https://github.com/sprice134/HUGO}{https://github.com/sprice134/HUGO}.


\end{abstract}

\section{Introduction}

Cold spraying is a solid-state additive manufacturing process that employs metallic powders as feedstock. These powders are propelled at supersonic velocities toward a substrate, where they undergo severe plastic deformation and subsequently bond to form material in a layer-by-layer fashion \cite{champagne2021practical}. In contrast to most additive manufacturing techniques, which depend on melting and solidification, cold spray remains entirely solid-state. As a result, cold spray is a highly versatile technology with applications spanning protective coatings, component restoration, and the fabrication of dense three-dimensional parts. This process also offers notable sustainability advantages, enabling the repair and life-extension of damaged components, thereby reducing material waste. Furthermore, overall process sustainability can be enhanced through generated or recycled carrier gases, as well as reclamation and reuse of unconsumed feedstock powder \cite{dowling2025feasibility,rua2025reusing,ozdemir2021high,champagne2016unique}.

While cold spray additive manufacturing offers numerous advantages, it is an inherently complex process influenced by a wide range of variables that must be carefully optimized to achieve desirable final part properties. Numerous experimental parameters can affect these properties, including carrier gas type, cold spray system design, powder composition, morphology, and treatment, deposition settings, substrate surface preparations, and others \cite{champagne2021practical}. Although prior studies have investigated the effects of individual cold spray processing parameters on part properties, relatively little work has examined how these interact within the broader processing framework. For example, Dowling et al. found that changing carrier gas purity affected one feedstock material but not another, highlighting this inter-dependent nature within process optimization \cite{dowling2025feasibility}. Similarly, studies isolating factors such as spray angle or powder morphology provide valuable conclusions, but are specialized and difficult to extend without a broader data foundation \cite{grubbs2023exploring}. Thus, multi-variable optimization is essential for achieving high-quality cold spray deposition, as modifying any single parameter often necessitates corresponding adjustments to others. Given this complexity and the wide range of tunable inputs, a comprehensive method for compiling and analyzing reported cold spray data would be highly valuable for process understanding, optimization, and modeling.

Prior works have curated small cold-spray data collections by manually extracting reported experimental values along with a  limited set of processing attributes from scientific literature \cite{huang2018post,wu2023significant,sinclair2021sintering,eftekhari2024comparative,wang2024towards,coddet2016mechanical,luo2014high,sample2020factors,sudigdo2025cold,wei2020solid,yin2024towards}. These extracted results, typically presented as summary tables, are quite small, frequently spanning only dozens of papers with the largest collection containing only 137 experiments. This small size limits their ability to serve as a training set for developing high-performing predictive models or for implementing effective process-optimization workflows. 

Additionally, while these collections may provide valuable preliminary insights, they tend to be  custom designed around a specific material class, processing variant, or target property \cite{sudigdo2025cold, luo2014high}, rather than aiming to capture a broad  general-purpose representation of cold-spray processing and outcomes. As a result, they often record only the particular subset of features needed for the targeted analysis. For example, Yin et al.~\cite{yin2024towards} focus on gas and peening-particle attributes for micro-forging-assisted cold spray, while Sinclair-Adamson et al.~\cite{sinclair2021sintering} collapse processing and post-treatment details for Al~6061 into a single free-text ``Method'' field, omitting factors such as powder morphology, substrate preparation, or standoff distance that have been shown to impact mechanical properties \cite{wong2013effect,kumar2016influence,li2008effect}.

To overcome these limitations, this work presents HUGO-CS, a large-scale dataset of experimental cold-spray mechanical-property results curated from scientific literature, containing 4,383 experiments with 144 features extracted from 1,124 sources. Creating a dataset that significantly exceeds prior works in both the number of experiments and the experimental parameters under consideration would be extremely time-consuming and costly if done entirely by hand. In this work, manually extracting information from a single article required an average of 91 minutes, yielding around  3.9 experiments per article.  In contrast, Large Language Models (LLMs) can extract information much more quickly. However, LLMs are prone to structural errors, omissions, and hallucinations when parsing complex inputs like scientific literature. Thus, to create this dataset at scale, this work introduces HUGO, a \textbf{H}ybrid-labeled, \textbf{U}ncertainty-aware, \textbf{G}eneral-purpose, \textbf{O}bservational extraction framework combining LLM extraction with targeted manual re-labeling to produce structured experimental records from primary sources.

In addition to scale, HUGO-CS is designed as a benchmark-ready dataset that can be used directly for meta-analysis or modeling. To achieve this, significant post-processing and data cleaning have been performed. In particular, free-text descriptors (e.g., feedstock and spray-systems names, carrier gas, and treatment type) are consolidated into standardized categorical values, feedstock chemistries are mapped to structured continuous composition records, and units are normalized across sources for consistent comparisons. Additionally, of the 4,383 extracted experiments, HUGO-CS contains a hand-labeled gold subset of 1,765 experiments from 243 sources, allowing users to work selectively with human-only labeled results if desired. Ultimately, HUGO-CS significantly expands the scale and accessibility of experimental cold-spray data, enabling modern data-driven analysis and predictive modeling that were previously constrained by small fragmented datasets.

In summary, this work offers the following contributions:
\begin{enumerate}
    \item \textbf{HUGO:} A novel hybrid-labeling framework that combines the efficiency of tailored LLM-based extraction with the accuracy of targeted manual verification, prioritizing articles most likely to contain extraction errors to maximize the value of additional manual labeling effort.
    \item \textbf{HUGO-CS:} A large-scale machine-readable dataset of \textbf{C}old-\textbf{S}pray mechanical properties, containing 4,383 experiments with 144 features per experiment across 1,124 primary sources, substantially exceeding the scale and feature coverage of prior cold-spray literature extractions.
    \item A human-labeled gold subset of HUGO-CS, consisting of 1,765 experiments from 243 sources, providing a high-fidelity reference set for benchmarking, estimating error rates, and analyzing failure modes.
    \item An extensive data cleaning and standardization pipeline that expands the usability of HUGO-CS for downstream tasks including model training, process optimization, and meta-reviews.
    \item A critical evaluation of LLM-based extraction performance on cold-spray literature, highlighting common failure modes and their impact on experimental coverage and label quality.
    \item Open access to the complete HUGO-CS dataset and the HUGO pipeline code to support extension to newly published cold-spray literature and adaptation to new application domains (\url{https://github.com/sprice134/HUGO}).
\end{enumerate}

\section{Related Works}
\subsection{Existing Datasets of Cold Spray Properties}
Several studies have curated cold-spray datasets by extracting reported results from the literature \cite{huang2018post,wu2023significant,sinclair2021sintering,eftekhari2024comparative,wang2024towards,coddet2016mechanical,luo2014high,sample2020factors,sudigdo2025cold,wei2020solid,yin2024towards}. However, these compilations are small and tailored to specialized domains or specific tasks. As shown in Table \ref{tab:sota_comparison}, these datasets contain, on average, 16.45 source articles, with a total of 45.18 experiments each with 7.82 features per instance. The largest of these datasets, released by Yin et al. \cite{yin2024towards} in 2024, contains 137 experiments extracted from 27 articles, with each instance characterized by 15 features. Unfortunately, %As one critical disadvantage,
these datasets are typically embedded in PDF tables/figures rather than a machine-readable file (e.g., .json, .csv, or .txt). As a result, this restricts their utility for downstream applications, requiring re-extracting and reformatting the reported values again before they could be used for subsequent analysis or modeling.

\begin{table}[!h]
\centering
\caption{Comparison of the proposed dataset, HUGO-CS, to existing cold-spray datasets curated from the literature, indicating whether results are reported numerically or in figures, separated into atomic attributes, are machine-readable, and if alloy compositions are provided, along with the number of articles, experiments, and features.}
\label{tab:sota_comparison}
\begin{adjustbox}{max width=\textwidth}
\begin{tabular}{l c c c c c c c}
\hline
\textbf{Article} &
\shortstack{\textbf{Numeric}\\\textbf{Values}} &
\shortstack{\textbf{Atomic}\\\textbf{Features}} &
\shortstack{\textbf{Machine}\\\textbf{Readable}} &
\shortstack{\textbf{Alloy}\\\textbf{Composition}} &
\shortstack{\textbf{Number of }\\\textbf{Articles}} &
\shortstack{\textbf{Number of }\\\textbf{Experiments}} &
\shortstack{\textbf{Number of }\\\textbf{Features}} \\
\hline
Huang et al. \cite{huang2018post} & & & & & 12 & 30 & 4  \\
Wu et al. \cite{wu2023significant} & & \checkmark & & & 20 & 61 & 4  \\
Sinclair-Adamson et al. \cite{sinclair2021sintering}  & \checkmark & & & & 8  & 17    & 4  \\
Eftekhari et al. \cite{eftekhari2024comparative} & \checkmark & & & & 9 & 23 & 10 \\
Wang et al. \cite{wang2024towards} & \checkmark & & & & 11 & 35 & 7  \\
Coddet et al. \cite{coddet2016mechanical} & \checkmark & & & & 15 & 28 & 6  \\
Luo et al. \cite{luo2014high} & \checkmark & & & & 23 & 26 & 4  \\
Sample et al. \cite{sample2020factors} & \checkmark & & & & 17 & 74 & 9  \\
Sudigo et al. \cite{sudigdo2025cold} & \checkmark & & & & 31 & 58 & 10 \\
Wei et al. \cite{wei2020solid} & \checkmark & \checkmark & & & 8  & 8 & 9  \\
Yin et al. \cite{yin2024towards} & \checkmark & \checkmark & & & 27 & 137 & 15 \\
\textbf{Ours} & \checkmark & \checkmark & \checkmark & \checkmark &  \textbf{1,124}  & \textbf{4,383} &  \textbf{144}  \\
\hline
\end{tabular}
\end{adjustbox}
\end{table}

In addition to limited scale, existing cold-spray datasets are typically tailored to specific material systems, processing variants, or narrow target properties. For example, Sudigdo et al.~\cite{sudigdo2025cold} compiled a literature review focused on cold spraying Ni-based superalloys, examining how processing conditions influence deposit porosity, Luo et al.~\cite{luo2014high} reviewed hardness outcomes in cold-sprayed deposits relative to their bulk counterparts, and Eftekhari et al. \cite{eftekhari2024comparative} evaluated the impact processing parameters on copper-based cold spray depositions.

While these studies provide valuable insights to their specific  applications, they tend to be narrowly focused on specific processing parameters and target variables. As a result, they often leave out the experimental provenance and report only the features required for the relevant analysis. Finally, many of these reviews report information using reductive feature representations, recording only a material's common designation (e.g., Al~6061, IN~718), even when the underlying chemical composition is reported in the initial source. Yet without such composition-level information, it is difficult to model the effects of alloying elements or implement transfer learning for chemically similar yet distinct materials.

\subsection{Data Extraction Pipelines from Primary Source Documents}

Outside of cold spraying or additive manufacturing, pipelines to convert structured archival documents like PDFs into machine-readable formats have been an ongoing area of research across applications  ranging from financial and legal/government documents to broader scientific topics \cite{beltagy2019scibert,xu2020layoutlm}. These approaches can generally be broken into three stages: (i) \textit{structure recovery}, capturing article metadata (title, authors, year, etc.), section layouts (introduction, methods, etc.), and references \cite{lopez2009grobid,tkaczyk2015cermine}; (ii) \textit{text-extraction}, capturing parsable text, or leveraging Optical Character Recognition (OCR) for scanned or image-based sources \cite{smith2007tesseract}; and (iii) \textit{rich data-extraction}, extracting information from tables where multi-axis spatial information is necessary for comprehension \cite{mehta2018camelot}. Given the wide variety of paper formats used, and the large amount of novel research and experimental discovery present, there has been a growing interest in PDF extraction for the scientific community \cite{bast2017benchmark,meuschke2023benchmark}.

However, once PDFs are converted into a machine-readable format, effective knowledge extraction remains challenging. Within the materials and chemistry domain, ChemDataExtractor \cite{swain2016chemdataextractor} was one of the first systems designed for knowledge extraction. To achieve this, ChemDataExtractor leverages chemistry-aware natural language processing (NLP) tools, such as rule-based grammars, to capture information from the main text, figure captions, and tables, producing structured chemical records. Building on this, BatteryDataExtractor \cite{huang2022batterydataextractor} followed a similar pipeline but replaced rule-based grammars with a transformer-based extraction model and expanded the scope from isolated entity-value pairs to linked material and property records, enabling more structured representations of experimental relationships.

With the advent of LLMs, the performance and accessibility of NLP methods have improved significantly. For example, Polak et al. \cite{polak2024extracting} proposed \textit{ChatExtract}, a prompt-engineering and follow-up verification framework for conversational LLM-based extraction, allowing for generalizable extraction of material property triplets (Material, Value, Unit). Similarly, Immanuel et al. \cite{immanuel2025enhancing} leveraged Gemini 2.5 Flash to extract six experimental attributes (alloy name, composition, yield strength, ultimate tensile strength, elongation, and additive manufacturing process) from materials literature, structuring them into standardized datasets for downstream analysis. In comparison to these prior attempts, HUGO targets full experimental extraction, including material compositions, powder morphologies, substrate conditions, spray properties, treatment conditions, and multiple target variables. To handle more complex data structures that cannot be verified as easily as a material triplet, HUGO introduces Hierarchical Risk Mitigation (HRM) to focus manual intervention on the more challenging extractions, balancing automated extraction efficiency while maintaining accurate extractions. 

\section{Methods}
 
To extract experimental results from literature, two primary methods were implemented: hand-labeling and LLM-labeling. While hand-labeling offers higher accuracy and reliability, it is significantly more time-consuming. In this work, labeling a document required an average of \textit{91 minutes} per document. Over the 1,124 articles in this study found to have experimental results, hand-labeling all of them would require approximately 1,704 hours (71 days). In contrast, an LLM labeled the same set of documents with an average of \textit{2.4 minutes}. In addition to the accelerated labeling from a one-to-one comparison, an LLM-based approach can be parallelized to further reduce labeling times with minimal changes. However, parallelizing manual labeling requires additional trained annotators and may cause reduced consistency across labeled outputs. To balance label quality with throughput, this work introduces HUGO, a hybrid-labeling workflow where an LLM produces an initial structured extraction and a Hierarchical Risk Mitigation (HRM) module flags high-risk outputs for additional manual review and correction, as shown in Fig.~\ref{fig:mainArch}.

\subsection{Text Extraction}
Traditional LLMs are trained on text-based token sequences, operating as text-to-text models \cite{xu2020layoutlm}. As a result, they are designed to consume unstructured text as an input and produce unstructured text as an output. However, the primary sources evaluated in this work are structured documents, containing heterogeneous layouts (e.g., single/double column) and diverse table designs (e.g., left/right aligned and varying cell bordering conditions) that cannot be directly consumed by a text-to-text model. Thus, MinerU~\cite{wang2024mineru} was used to convert each structured PDF into a standardized, machine-readable markdown file (.md), preserving document structure (e.g., tables, captions, and sections) and providing a consistent form for scientific notation that can be lost when represented as free-text. 

\subsection{Metadata and Traceability} \label{sec:metadata_traceability}
For each article, the title was parsed from extracted text and the Crossref API \cite{crossref_rest_api} was used to retrieve associated metadata, including authors, journal, publisher, publication date, and DOI. Using string-similarity, articles with a single high-confidence title match were automatically linked to their Crossref record. However, when no strong match was found, often due to special characters being parsed in \LaTeX\ form by MinerU \cite{wang2024mineru}, or when multiple records had near-identical titles, the metadata was retrieved manually. In cases where multiple matches were identified, the original article was consulted and the title, venue (journal or proceedings), author list, and publication year were used for entity resolution. In cases where there was no DOI (e.g., datasheets, government reports, or conference proceedings), a non-DOI source link was recorded to preserve traceability to the original document. For example, in Champagne et al. \cite{champagne2008magnesium}, there was no available DOI, so the source link was recorded as the archived Defense Technical Information Center (DTIC) report. However, of the articles included in the HUGO-CS dataset, only 15 required a non-DOI source link, with nearly 99\% of entries containing an accessible DOI.

\subsection{Experimental Schema Construction}
Compared to prior studies, this work proposes a significantly larger schema (the set of extracted features) spanning material properties, experimental parameters, and testing conditions to represent a cold spray experiment. During construction, a multi-stage iterative approach was used to ensure broad coverage and maximize experimental representation. First, a small subset of candidate articles was probed to identify the experimental parameters modified and reported as a baseline. These candidate fields were then reviewed and consolidated by a domain expert to define the scope of the schema and add additional attributes when necessary. Thereafter, a complete review of the extracted dataset on the initial schema revealed missed attributes, such as process-specific parameters for Hot Isostatic Pressing (HIP) \cite{chen2019effect} or Laser-Assisted Cold Spraying (LA-CS) \cite{olakanmi2014laser}, that were absent from the initial subset, which were subsequently added to the schema. Finally, the  dataset was re-collected producing a final dataset with the
instances attributed with  the full 144 features.

\subsubsection{Experimental Inclusion and Exclusion Criteria}
Candidate articles for review were identified using SCOPUS \cite{baas2020scopus} with a keyword search targeting cold-spray studies that report mechanical properties (e.g., hardness, yield strength, ultimate tensile strength, elastic modulus, or elongation). However, this identified many false positives, including papers that mentioned cold spray only in related work, reported microstructure/process characterization without mechanical testing, or focused on other deposition processes (e.g., HVOF) while using cold spray only as background or comparison. Thus, this work established the following inclusion criteria. For an experiment to be included, it must (i) correspond to a cold-sprayed deposit specimen (not a wrought or bulk baseline, and not substrate-only results), (ii) represent a distinct set of processing and testing conditions, and (iii) report at least one target outcome (porosity, elastic modulus, yield strength, ultimate tensile strength, elongation-at-break, microhardness, or nanohardness). For a candidate article from literature to be included in the final dataset, it must report at least one eligible cold-spray experiment.
Additionally, to prevent duplicate reporting, experiments must be novel to that source to be included. Experimental results that were a summary of, or a reference to, prior experiments without new cold-spray measurements were excluded.

\subsection{Prompt Construction}
In this work, rather than training or fine-tuning a model, extraction was performed using instruction-based prompting with a general-purpose LLM. Prior works have shown that providing example input-output pairs, known as \textit{few-shot} or \textit{many-shot} prompting, can improve extraction performance \cite{brown2020language}. However, prompt-based methods are constrained by input length. Models have a maximum context length, where larger inputs cannot be processed. Additionally, even when inputs fall within the stated limit, models frequently have a significantly smaller \textit{effective} context length, with performance degrading as input size exceeds this lower limit~\cite{hsieh2024ruler,liu2024lost}. Considering this effective context length, the length of scientific articles, and the diverse locations of necessary information throughout, HUGO used \textit{zero-shot} inference, where no complete source article and extraction were provided as a sample. Instead, a detailed schema definition, inclusion/exclusion criteria, and an abridged example of a correctly extracted experiment were included. In addition, traditional prompt-engineering strategies were used, including model role assignment, a pre-prompt, explicit domain context, extraction rules, structured output formatting, an example output, and a final re-statement of the task \cite{reynolds2021prompt}.

\subsection{Experimental Extraction: LLM Labeling}
After defining the extraction schema and prompt, GPT o4-mini \cite{openai_o4_mini} was applied to generate an initial extraction for each article. This model was implemented with the OpenAI API, using public weights on June 10, 2025. LLM-based extraction was performed with \texttt{reasoning\_effort} set to \textit{High} (out of \textit{Low}/\textit{Medium}/\textit{High}). While \texttt{temperature} is frequently tuned for LLM research to constrain output variability, due to the reasoning component, \texttt{temperature} cannot be configured for the o4-mini model. This initial LLM-inference was applied to a total of 3,347 articles. Of these, 1,074 source articles were found to contain novel experimental data by the LLM, capturing 3,366 experiments, as shown in Table~\ref{tab:curation_summary}.

\subsection{Hierarchical Risk Mitigation with Manual Labeling} \label{sec:HRM_Main}
\begin{figure}[!h]
    \centering
    \includegraphics[width=0.65\linewidth]{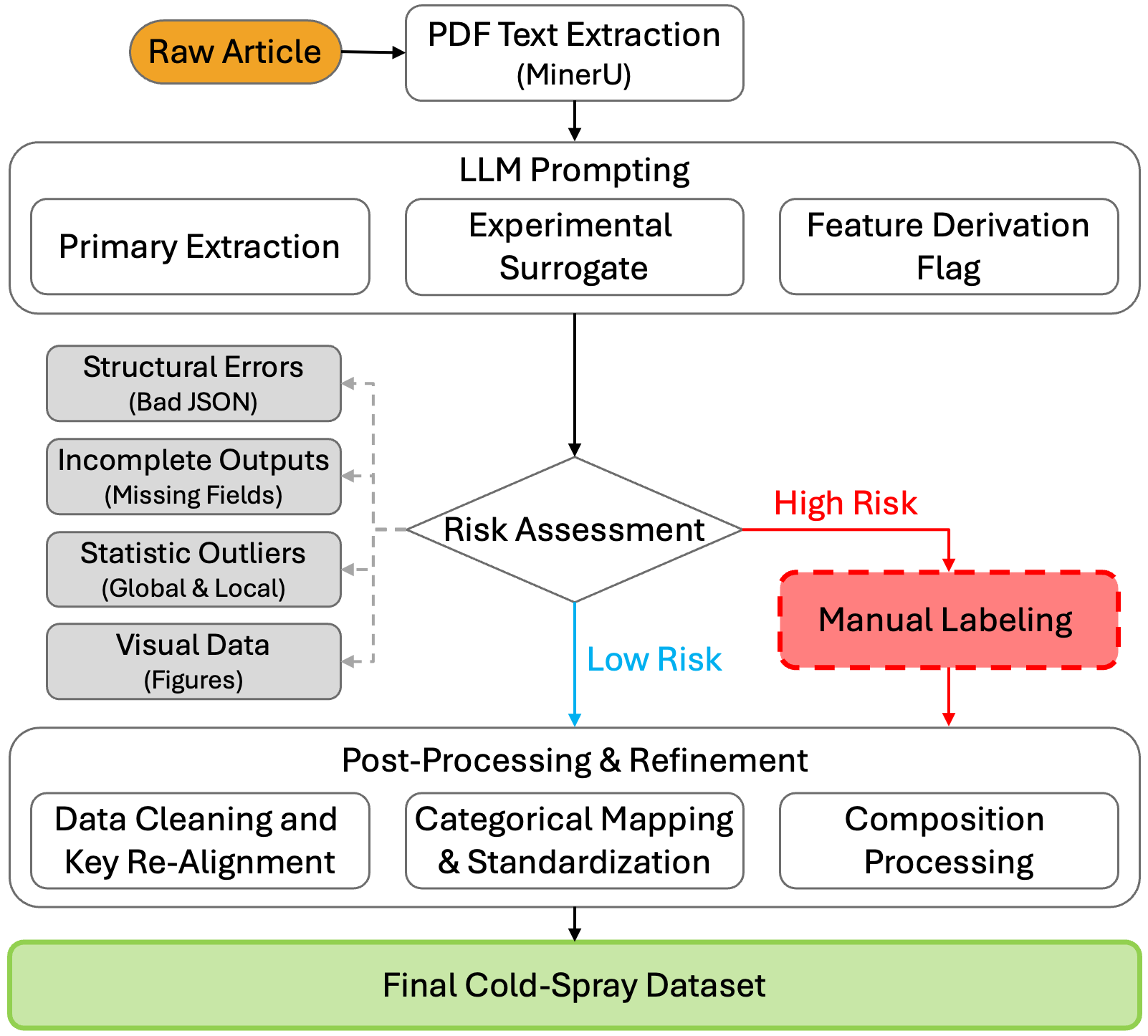}
    \caption{Overview of the HUGO pipeline for cold-spray dataset construction. Articles are converted to text (MinerU) and processed by an LLM to create an initial labeling. Hierarchical Risk Mitigation (HRM) identifies and flags experiments with a high risk for error to be manually labeled. Finally, experiments (LLM-labeled and hand-labeled) are post-processed, including categorical string mappings, continuous mapping for compositions, and unit standardization to produce HUGO-CS, the final cold spray dataset.}
    \label{fig:mainArch}
\end{figure}

One of the primary contributions of the HUGO framework is the Hierarchical Risk Mitigation (HRM) module, a targeted manual-labeling strategy tailored to the specific failure modes of LLM-based experimental extraction, as shown in Fig.~\ref{fig:mainArch}. Rather than a uniform approach that treats all sources as equal and randomly labels a subset of articles, HRM prioritizes documents whose LLM outputs exhibit the highest risk of error. Articles with outputs that are structurally valid, schema-aligned, and numerically consistent are deemed acceptable and kept. This results in a cost-effective correction process, maximizing the impact of manual intervention while preserving the efficiency of automated labeling. To achieve this hierarchical approach, the HRM flags documents for manual labeling in multiple stages, as defined below. 

\subsubsection{Structural Errors: Unparsable JSON Extractions}\label{sec:HRM_Stage1}
As shown in Fig.~\ref{fig:mainArch}, the first stage of HRM flags structurally invalid LLM outputs for manual labeling. In this work, the LLM was instructed to extract experimental results from the document and report them into a structured JSON form. While successful in most cases, as a probabilistic model, this structured formatting is not guaranteed, even when paired with schema-specific decoding modules \cite{geng2025generating}. In an automated dataset-curation workflow, a strictly structured output is required for downstream parsing and validation. When the output is malformed or truncated, additional re-extraction or repair steps become necessary, increasing pipeline complexity and introducing additional sources of error. As a result, the first set of articles flagged by HRM were articles where the LLM failed to produce a valid, machine-readable output. These unparsable outputs occurred most frequently among the longest responses, containing the most extracted experiments. For example, some outputs were syntactically invalid (e.g., incorrectly ordered braces or missing delimiters), some exceeded the maximum output length and were truncated mid-response, and others were conversational, such as ``\textit{I'd be happy to.}'', without actually providing the extraction.

\subsubsection{Incomplete Outputs: Incorrect Schema Extractions}\label{sec:HRM_Stage2}

In some cases, the LLM produced a valid JSON extraction that could be parsed, but failed to adhere to the desired schema for extracted experiments, dropping desired features or introducing new ones \cite{tang2024struc}. Thus, as shown in Fig.~\ref{fig:mainArch}, the second stage of HRM targeted these schema-compliance errors from the remaining extractions. For missing features, this could reflect either a case where the source article did not discuss that attribute and, rather than returning an empty value, the LLM omitted the field, or a more significant extraction error where the article reported the information but the LLM failed to capture it. Similarly, unexpected additional fields could reflect either undesired information outside the scope of the current extraction, or desired information that the LLM correctly extracted but returned under an incorrect feature name.

All articles containing at least one experiment with a non-compliant schema were initially flagged for manual re-labeling. However, to reduce the burden on manual labeling, two assumptions were made for specific failure cases where automated correction could be performed. First, for several optional sub-processes, the LLM was instructed to extract both a binary indicator of whether the sub-process occurred and the associated parameter fields. For example, in the case of micro-forging, this included a binary flag indicating whether micro-forging was used, along with attributes describing the size, shape, and volume fraction of the forging particles. In cases where the binary flag was extracted as \texttt{false}, but the corresponding parameter fields were omitted, those omitted fields were assumed to be empty and filled accordingly.

Alternatively, in cases where an extracted experiment contained both omitted features \textit{and} unexpected additional features, string-similarity matching was used to determine whether the unexpected fields corresponded to the missing features but were named incorrectly, as shown in Table~\ref{tab:llm_similarity}. Each unexpected feature was compared against the set of omitted features, and if the best-match string similarity was $\geq$ 0.8, the unexpected feature was reassigned to the corresponding missing attribute. A threshold of 0.8 was chosen as a conservative estimate, where manual inspection found no incorrect mappings over this value. If neither of these conditions was met, or if the experiment still failed to adhere to the desired schema after these correction steps, the article was re-labeled by hand.

\begin{table}[!h]
\centering
\caption{Examples of LLM-produced schema errors (red) and their corrected forms (green), using string-similarity scores for automated correction.}
\label{tab:llm_similarity}
\begin{tabular}{l | l | c}
\hline
\textbf{LLM Output} & \textbf{Correct Value} & \textbf{Similarity} \\
\hline
Majority\_Powder\_Primary\_\hlr{ELEMENT} &
Majority\_Powder\_Primary\_\hlg{Element} & 1.00 \\

Laser\_\hlr{PWR}\_Value &
Laser\_\hlg{Power}\_Value & 0.94 \\

Hot\_Rolling\_Reduction\_Ratio\_Unit\hlr{ } &
Hot\_Rolling\_Reduction\_Ratio\_Unit\hlg{s} & 0.98 \\

Powder\_Particle\_Mean\_\hlr{SIZE}\_Value &
Powder\_Particle\_Mean\_\hlg{Size}\_Value & 1.00 \\

Deposit\hlr{-}Microhardness\_System &
Deposit\hlg{\_}Microhardness\_System & 0.98 \\

\hlr{ }Microhardness\_Value &
\hlg{Deposit\_}Microhardness\_Value & 0.83 \\
\hline
\end{tabular}
\end{table}

\subsubsection{Statistical Outliers: Content-Based Anomalies}
The first two stages of HRM focus on syntax, identifying outputs that are either unparsable or non-compliant with the desired schema. However, these checks remain largely content-agnostic, and, therefore cannot detect cases where an extraction is structurally valid but contains incorrect values. Examples of such failure cases include errors introduced during PDF-to-text conversion (MinerU parsing), unit inconsistencies, or logical errors by the LLM during extraction. Thus, as shown in Fig.~\ref{fig:mainArch}, the third stage of HRM introduces content-based screening to identify statistically anomalous values in syntactically correct extractions. This was performed in three passes: domain-informed thresholds, global outliers, and local outliers.

First, domain-informed thresholds were used to specify physically reasonable limits (e.g., porosity values approaching 100\%, elastic modulus values near zero, or strengths orders of magnitude outside expected ranges). While effective for the most egregious errors, these filters are limited by the choice of appropriate thresholds. Thus, additional outlier detection was performed at the global and local levels to identify errors not captured by fixed domain thresholds. In this work, global outliers were values that were statistically anomalous to the full dataset, while local outliers were values that were anomalous within their specific alloy class. For global outliers, any experiment whose target feature deviated by more than $3\sigma$ from the dataset mean for that property was flagged for manual re-labeling. For example, if an experiment had an unusually high porosity relative to the overall dataset but had not been flagged by the domain threshold, it could still be flagged. For local outliers, due to data sparsity and inconsistent reporting of target features across studies, each extracted target value was first normalized to a standard unit convention (Sec.~\ref{sec:postProcessing}) and represented in a shared vector space. The distance of each experiment from the centroid of its corresponding material class was then compared against the class spread, and experiments exceeding with distance greater than $2\sigma$ were flagged for re-labeling.

Unlike the first two stages of HRM, which identify experiments with definite extraction failures (e.g., unparsable outputs or incorrect schemas), outlier detection identifies experiments that are more likely to contain an error, but are not guaranteed to be incorrect. As discussed further in Sec.~\ref{sec:ResultsOutliers}, some flagged values were verified against the source article and found to be correctly extracted despite being anomalous. In these cases, the flagged value was inspected but left unchanged. Since this review was limited to the outlier verification rather than the full article, the corresponding experiment remained marked as LLM-labeled in final dataset.

\subsubsection{Coverage Errors: Results Reported Only in Figures}
As shown in Fig.~\ref{fig:mainArch}, the fourth stage of HRM targets coverage errors, aiming to recover experiments that were missed by the LLM during the extraction process. In many cases, this occurred because results were reported in figures rather than in the machine-readable text used for extraction. Image-based extraction methods were explored, but produced inconsistent results that were difficult to align reliably with the corresponding experimental records extracted from text. In particular, figure extraction requires correctly interpreting titles, axis labels and scales, multi-axis plots (e.g., yield strength on one axis and elastic modulus on the other), color-coded experimental conditions, and other visual structure. Expanding HUGO to support image-based extraction will be explored in future work, but the current framework is limited to text-based information.

To address this coverage gap and identify LLM-labels with a high likelihood of omitted experiments, a second targeted LLM prompt was used to generate an estimate of the expected number of experiments for each document. As scientific, peer-reviewed articles, these sources frequently carefully defined the experimental design in the methodology section, including the number of samples created and types of characterization performed. Thus, rather than prompting the model to extract the experimental values themselves, this secondary prompt instructed the LLM to identify how many experiments were conducted and which target measurements were reported based on the described methodology and experimental design. Compared to extraction, this was not dependent on if these values were present in the text. Additionally, this secondary extraction is a simpler task than full experimental extraction, and could still succeed when the primary extraction failed to recover all experiments. For example, if an article reported 30 experiments, capturing all 144 features for each experiment could likely result in missing experiments or abridged outputs if the model timed out. However, simply identifying that 30 experiments were performed is less likely to produce a similar issue. Articles with the largest gap between the number of extracted experiments, and number of expected experiments from the secondary prompting were prioritized for manual labeling, where the expected coverage gains were highest.

To achieve this, each article was assigned an \textit{Expected Value} (\textit{EV}), representing the number of additional experiments expected to be gained through manual labeling. For example, if the secondary prompt estimated that an article contained eight experiments, but the extracted results contained only three, then the article would have an \textit{EV} of five. In cases where only a subset of target metrics was extracted (e.g., only yield strength and ultimate tensile strength), the \textit{EV} was prorated according to the fraction of expected target variables that were missing, balancing the number of distinct experiments with completeness over the desired target features. 

As shown in Fig.~\ref{fig:subA_features}, the frequency that each target metric is reported varies significantly across papers, ranging from 506 extracted yield strength values to 2,980 extracted microhardness values. To prioritize coverage gains for these sparser metrics, a \textit{weighted EV} (\textit{wEV}) was computed, weighting each target metric by the inverse frequency. As a result, if two articles have identical EV scores, but one is expected to contain mostly missing yield strength values, while the other is mostly missing microhardness values, the \textit{wEV} would be higher for the yield strength article. In this work, a threshold of $T=2.5$ was used, where any article with \textit{wEV} $\geq T$ was flagged for hand-labeling. However, $T$ can be dynamically set based on available labeling resources and target objectives: larger values reduce labeling costs but capture fewer articles, while smaller values increase coverage at the expense of additional manual labeling.

\subsection{LLM Extraction Performance Evaluation}
Of the articles hand-labeled in this study, many were selected by the HRM because they were considered high risk for extraction errors. As a result, evaluating LLM performance only on this HRM-flagged subset would likely overestimate the error rate of the remaining LLM-labeled articles, since this subset was intentionally selected as difficult examples where the LLM likely failed. Thus, to obtain a more representative estimate of performance on the non-flagged LLM-labeled samples, a held-out evaluation set was constructed by randomly sampling from the remaining sources and manually labeling them. For this held-out evaluation set, 20 articles were randomly selected from the non-flagged sources for manual labeling, yielding a total of 80 ground-truth experiments for evaluation. This held-out set, while randomly selected, still provided coverage across all target variables and a variety of possible extraction conditions. In particular, this set included articles with single- and double-column layouts, results reported in main text, tables, and figures, as well as cases where experimental values were repeated from prior sources and therefore did not satisfy the inclusion criteria.

Even when the ground truth labels existing, computing accuracy for an LLM-labeled article is not a trivial task. One of the key challenges is the ambiguity of ordering. That is, even when evaluation is restricted to a single source, both the LLM-labeled and hand-labeled outputs are \textit{unordered} sets of experimental records. As a result, experiment ordering is not guaranteed to be consistent across labeling methods and direct index-wise comparison may incorrectly treat differently ordered lists as incorrect. Thus, for each article, a pairwise similarity matrix between all ground-truth experiments and all LLM experiments was constructed. Each similarity is computed as the average per-key value similarity, using relative error for numeric fields and string similarity for text-based fields. Hungarian assignment \cite{kuhn1955hungarian} was then applied to the matrix to compute the optimal one-to-one experimental matching between LLM- and hand-labeled extractions.

%article 103 has two columns, reports hardness and porosity in text
%article 2191 has one column, reports hardness
%ARTICLE 1480 has two column, reports modulus and uts
%ARTICLE 825 has two column, hardness and porosity, and information in both text and figures

\subsection{Post-Processing \& Data Cleaning} \label{sec:postProcessing}

Using the proposed HUGO framework, this work curated one of the largest datasets of cold spray experimental results to date, titled \textit{HUGO-CS}. However, in its raw form, this dataset would be difficult to use for literature review, meta-analysis, or modeling, because many attributes are stored as free-text values. When stored as text, identical values can be expressed in many different forms. For example, the same feedstock may be reported as ``AA~6061'', ``Al~6061'', ``Aluminum 6061'', ``6061 Al'', or ``Al~6061 powder'', depending on the author. During the labeling process, both LLM and manual approaches captured these values as reported, resulting in 19 unique values describing \textit{316L Stainless Steel} powder feedstock, 15 describing \textit{Grade 2 Titanium}, and 14 describing \textit{Aluminum 6061}. Similar 
linguistic paraphrasing was observed across a variety of attributes, such as carrier gas (e.g., ``N2'' vs.\ ``$N_2$'' vs.\ ``Nitrogen'') or spray system (e.g., ``VRC Gen III'' vs.\ ``VRC Gen 3'' vs.\ ``Gen III Cold Spray System (VRC)''). Without consolidation, these diverse representations
fragment the dataset into many semantically identical categories, artificially increasing dataset sparsity and long-tail class distributions. 

To address these limitations and accelerate downstream tasks, additional post-processing steps were created to clean extracted records, including consolidating free-text attributes. Namely, 
within the HUGO framework, this work introduces an additional module, called
Propose-Inspect-Review (PIR), a second hybrid approach for dataset curation. Namely, using a combination of algorithmic and LLM-based methods, candidate string mappings are proposed (e.g., ``Aluminum 6061 Powder'' $\rightarrow$ ``Al~6061''). A user can then inspect these candidates, accept the desired mappings and prune the rest. Finally, an LLM performs a lightweight review pass to flag potentially conflicting logic, overlapping values, or mapping mistakes. After an initial pass, this  PIR workflow can be applied iteratively, narrowing the remaining alias space and providing additional examples of the desired mapping style and level of granularity to guide subsequent proposals.

\subsubsection{Categorical String Processing}
For categorical features, this was a relatively straightforward process. String-similarity matching and LLM inference were used to propose candidate mappings, and many aliases could be consolidated with minimal manual intervention. However, in some cases, domain knowledge was necessary to correctly merge terminology and avoid incorrect equivalences. For example, while Ti-6Al-4V and Grade~23 Ti-6Al-4V are closely related, Grade~23 has stricter purity requirements and should not be merged with standard Ti-6Al-4V. In contrast, common aliases for Ti-6Al-4V include Ti64, Grade~5 titanium, and TC4 (commonly used in China). By including these post-processing steps in the HUGO-CS dataset, downstream applications do not require as significant of domain knowledge for data cleaning and users can more readily filter, merge, and compare experiments across sources using standardized categorical representations.

\subsubsection{Continuous String Processing}
For continuous strings, such as powder feedstock composition, a similar PIR workflow was designed. However, unlike categorical attributes, these strings do not cleanly map to a single label because they encode multi-element mixtures with varying proportions. This continuous representation enables richer analysis that is not accessible from a categorical name alone. For example, compositional similarity can be compared across samples or, models can learn trends in alloy behavior (e.g., differences among Al-based alloys or the effect of chromium content on mechanical properties). Thus, in this work, powder feedstock compositions were mapped to a 50-element representation derived from the set of elements observed in the reported HUGO-CS compositions, assigning each element its percent composition. However, this composition mapping process was complicated by three challenges: diverse reporting styles, inconsistent units, and powder blends. 

\begin{figure}[!h]
    \centering
    \includegraphics[width=0.8\textwidth]{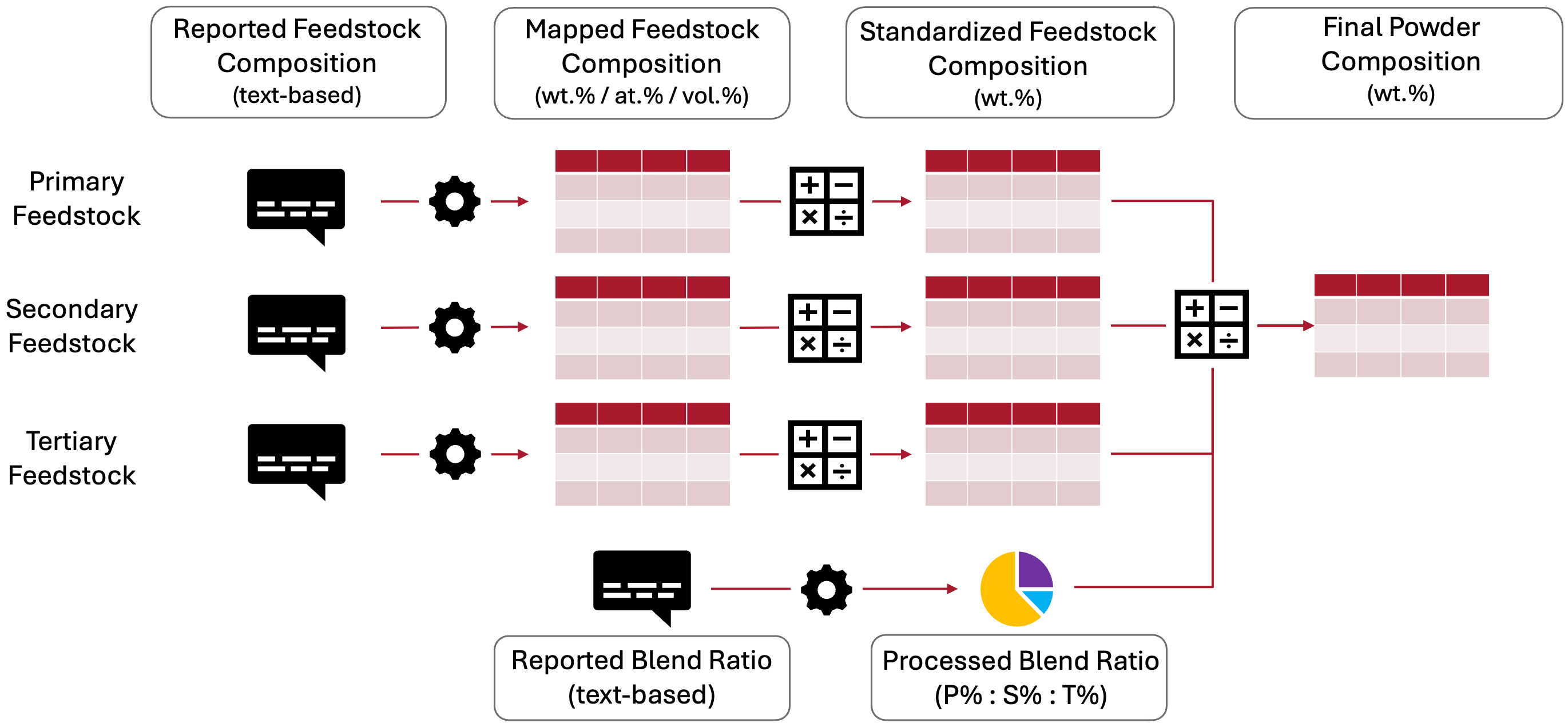}
    \caption{Visualization of composition processing in the HUGO framework. For each experiment, the primary, secondary, and tertiary feedstock are captured (when available) as free-text. These text-based values are mapped to a 50-element structured set in their native units (e.g., wt.\%, at.\%, or vol.\%), and then normalized to wt.\% using known elemental values. In cases of blends, powder blend ratios are captured as free-text and converted into  numeric ratios to compute a final composition in wt.\%.}
    \label{fig:CompositionWorkflow}
\end{figure}

To address each of these, a multi-stage composition processing approach was designed, as shown in Fig.~\ref{fig:CompositionWorkflow}. Specifically, to address diverse reporting styles, LLM-based proposals were used to extract a minimal composition structure consisting of element-value pairs  reported. These pairs were then inspected by hand and, when necessary, corrected. Once confirmed, an algorithmic process expanded this minimal structure into the full 50-element schema by setting missing elements to 0 and computing the balance composition, when present. In the review stage, an LLM was used to flag potential logical inconsistencies (e.g., impossible element sets or malformed ranges), while an algorithmic check verified that compositions summed to 100\%. To handle inconsistent units, compositions were initially preserved in their reported form (wt.\%, at.\%, or vol.\%), and values reported in at.\% or vol.\% were later algorithmically converted to wt.\%. For powder blends, HUGO extracted the primary, secondary, and tertiary feedstock compositions as well as the reported mixing ratio (when available). Each feedstock component composition was processed independently through the composition-mapping workflow to produce normalized wt.\% compositions. In parallel, a separate PIR pass standardized the free-text blend description into a numeric mixing ratio. Finally, an aggregation step was performed to compute the blended wt.\% composition.

\subsubsection{Unit Normalization}
After consolidating categorical values and mapping free-text continuous fields into structured representations, numeric units were standardized to improve comparability. Using an automated value-unit parsing module, reported units were normalized into a standard form (e.g., MPa for strength, GPa for elastic modulus, mm and \SI{}{\micro\metre} for length/size statistics, \SI{}{\degreeCelsius} for temperature, L/min for gas flow, mm/s for traverse speed, and degrees for angles). Partially non-numeric results were converted into purely numeric values. For example, inequalities (e.g., $\ge$1\%'' porosity) were reduced to their reported bound, and numeric ranges (e.g., ``1--5\%'') were converted to their mean. 

However, hardness required additional domain-specific handling. Namely, in addition to different notations, such as HV, HRB, or GPa, hardness units can encode the testing process used. For example, Vickers hardness is often reported with an appended load indicator (e.g., HV0.3 or HV100), where the numeric suffix denotes the applied test load. In these cases, the reported unit was normalized to ``HV'' and the suffix value was parsed as the hardness test load. When the extracted test-load field was empty, the parsed suffix was used to populate that field to preserve experimental provenance. Values with incompatible, missing, or unrecognized units were flagged as \texttt{invalid}, allowing easier downstream filtering. Additionally, while these conversions improve utility for applications like modeling, they can result in a loss of information, such as converting a range to a single value. As a result, the standardized units are stored as a secondary dataset in the included GitHub repository, allowing users to access the original information, if desired.

\subsection{Secondary Labeling for Improved Experimental Provenance}\label{sec:YS_Relabel}
One challenge in constructing a standardized machine-readable dataset from heterogeneous literature is aligning diverse experimental procedures into a consistent representation. A concrete example of this is \textit{yield strength} and \textit{elastic modulus} as target features. Depending on the procedure used in these values, measurements can reflect fundamentally different property definitions and levels of comparability. For example, in cold spraying, many studies follow ASTM E8 \cite{ASTM_E8E8M_25} uniaxial tensile testing to report \textit{yield strength} and \textit{elastic modulus}. However, uniaxial testing often requires large deposit volumes and specimen machining, which can be cost-prohibitive. 

As a result, some articles report \textit{yield strength} and \textit{elastic modulus} estimates derived from nanoindentation \cite{campbell2020investigation,tsaknopoulos2022through}, or profilometry-based indentation plastometry (PIP) \cite{lama2024role,tsaknopoulos2022evaluation}, which can produce values that are similar but not directly comparable to ASTM E8 tensile measurements \cite{bhat2020constraint}. Naively removing these values would substantially reduce coverage. However, retaining them without additional provenance labels could negatively impact downstream analysis and model training. To address this, a secondary document-level labeling pass extracted if mechanical properties were derived from non-standard test methods and recorded both the affected properties and the experimental procedure used. 

\section{Results}

\subsection{Snapshot of HUGO-CS Dataset}
\begin{figure}[!h]
  \centering
  \begin{subfigure}[t]{0.49\textwidth}
    \centering
    \includegraphics[width=\linewidth]{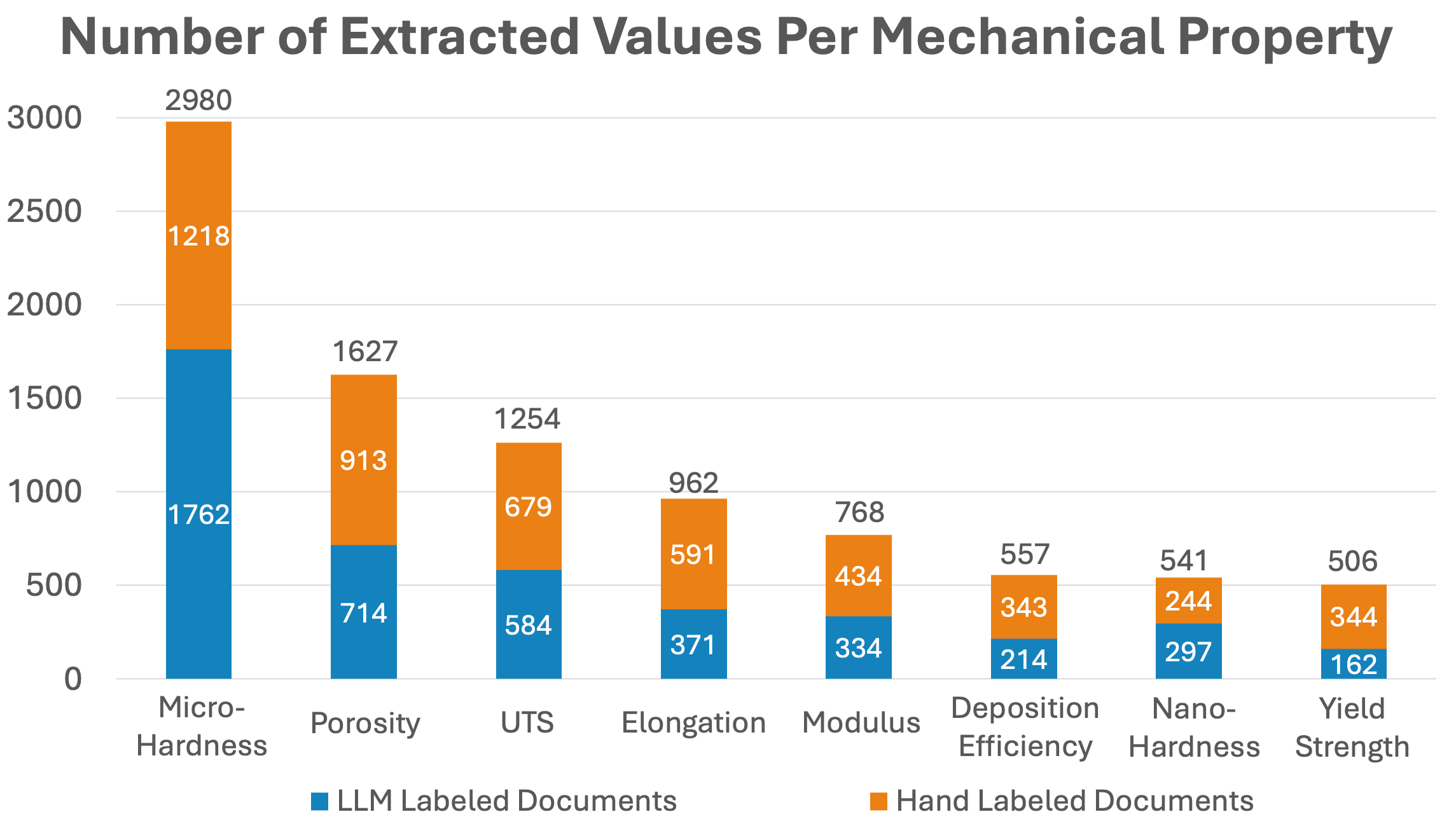}
    \caption{}
    \label{fig:subA_features}
  \end{subfigure}\hfill
  \begin{subfigure}[t]{0.49\textwidth}
    \centering
    \includegraphics[width=\linewidth]{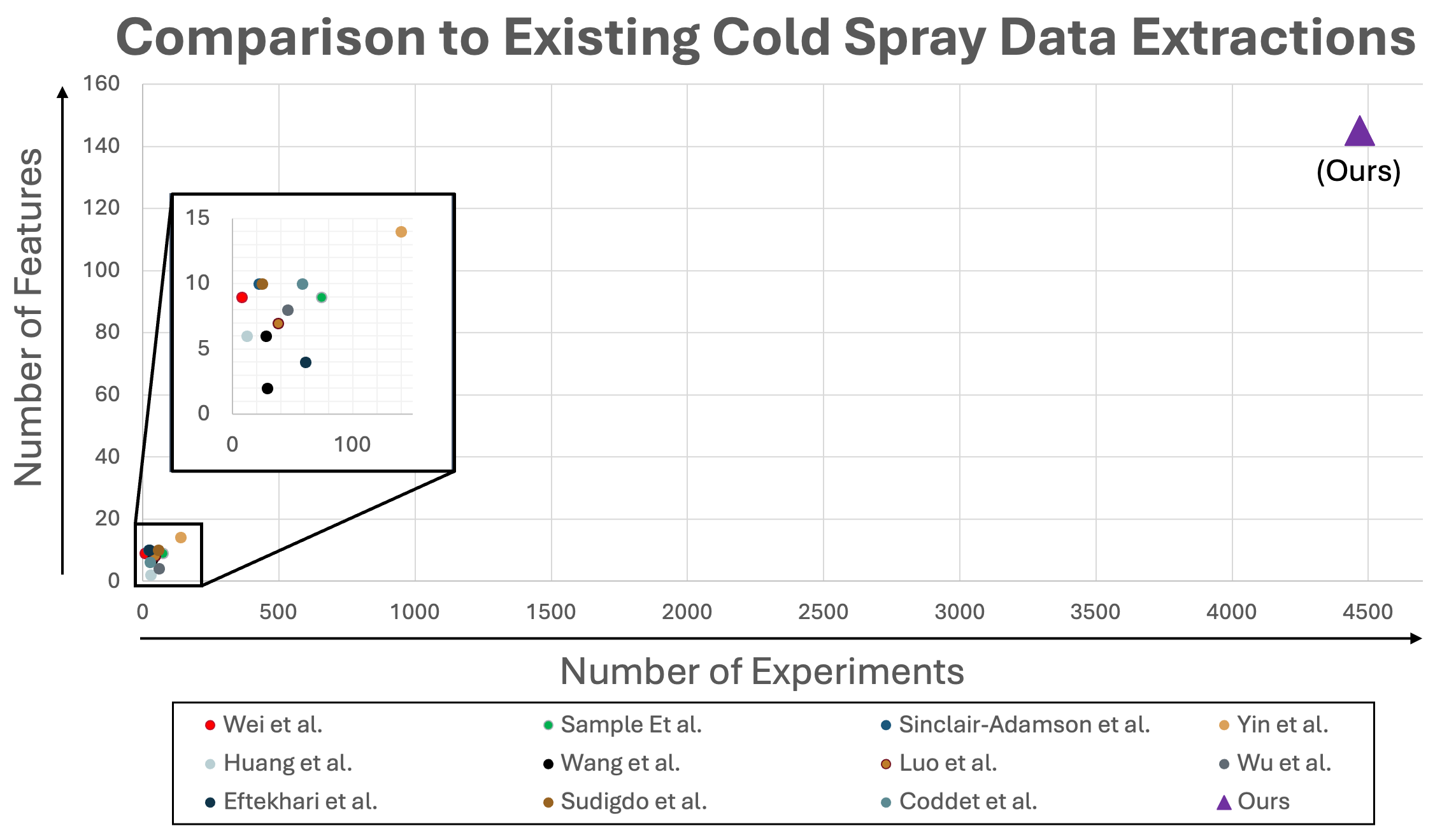}
    \caption{}
    \label{fig:subB_sota}
  \end{subfigure}
  \caption{(A) Visualization of extracted values per mechanical property, where blue indicates LLM-extracted results and orange indicates manually labeled values. (B) Comparison of the number of features and experiments against existing cold spray datasets, highlighting the increased scale and granularity of the dataset introduced in this work over the state-of-the-art.}
  \label{fig:sizeDemonstration}
\end{figure}

\noindent\textbf{Hybrid Labeled.} After all labeling (LLM + manual) was completed, the hybrid-labeled \textbf{H}UGO-CS dataset contained 4,383 experiments across 1,124 articles, with a hand-labeled gold subset of 1,765 experiments across 225 articles (and an additional 18 manually labeled articles found to contain zero novel cold-spray experimental results). Broken down by target variable (Fig.~\ref{fig:subA_features}), microhardness was the most frequently reported property with 2,980 values, while yield strength was the least with 506 values. For each target variable, at least 41\% of values were labeled by hand, with a maximum of 68\% for the yield strength property. In comparison to prior cold-spray dataset curations (Fig.~\ref{fig:subB_sota} and Table~\ref{tab:sota_comparison}), this work significant increases the number of experiments and the granularity of process-parameter coverage  available to researchers and practitioners to date in the cold spray field.

\begin{figure}[t]
  \centering
  \begin{subfigure}[t]{0.49\textwidth}
    \centering
    \includegraphics[width=\linewidth]{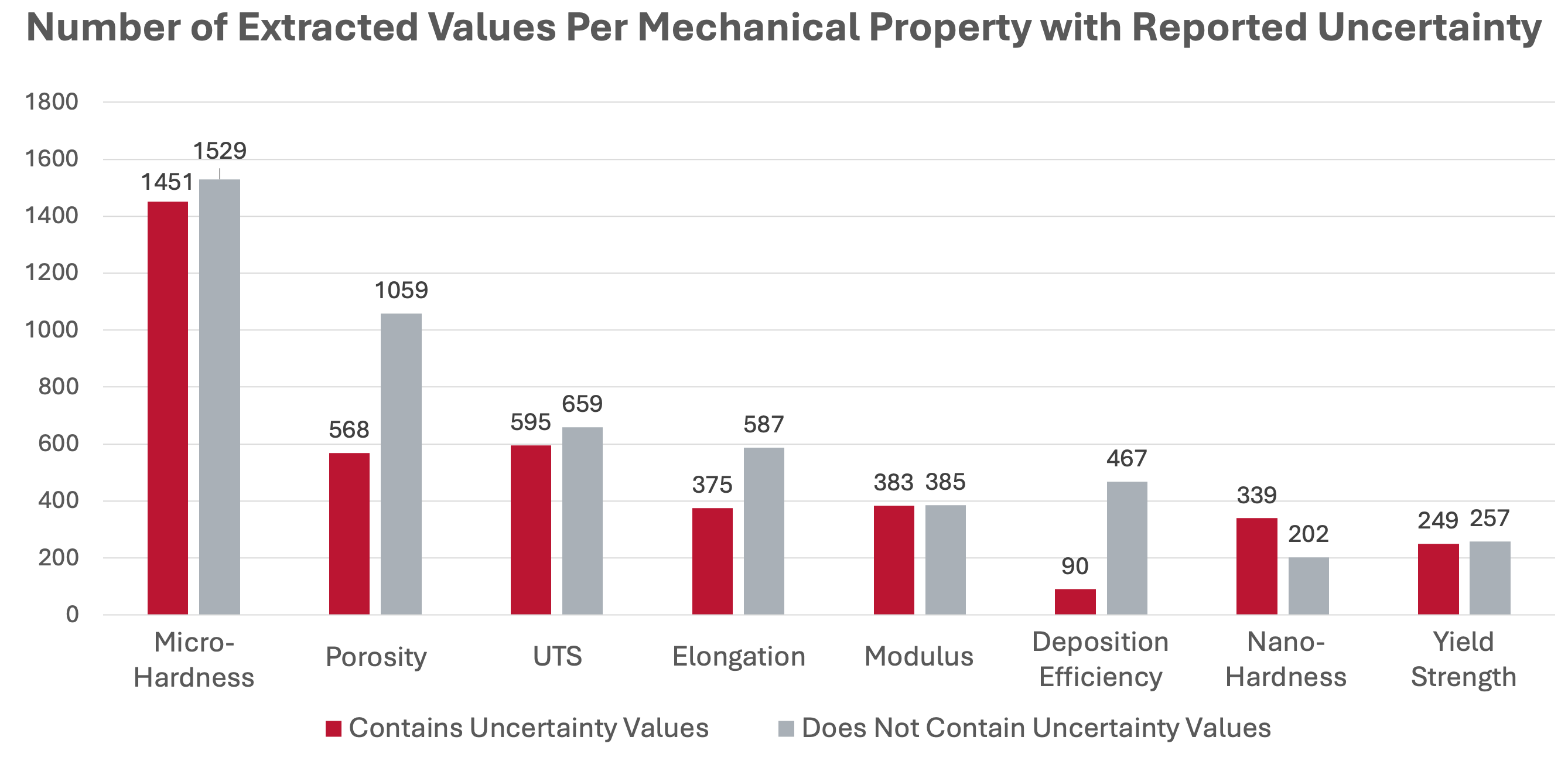}
    \caption{}
    \label{fig:uncertainty}
  \end{subfigure}\hfill
  \begin{subfigure}[t]{0.49\textwidth}
    \centering
    \includegraphics[width=\linewidth]{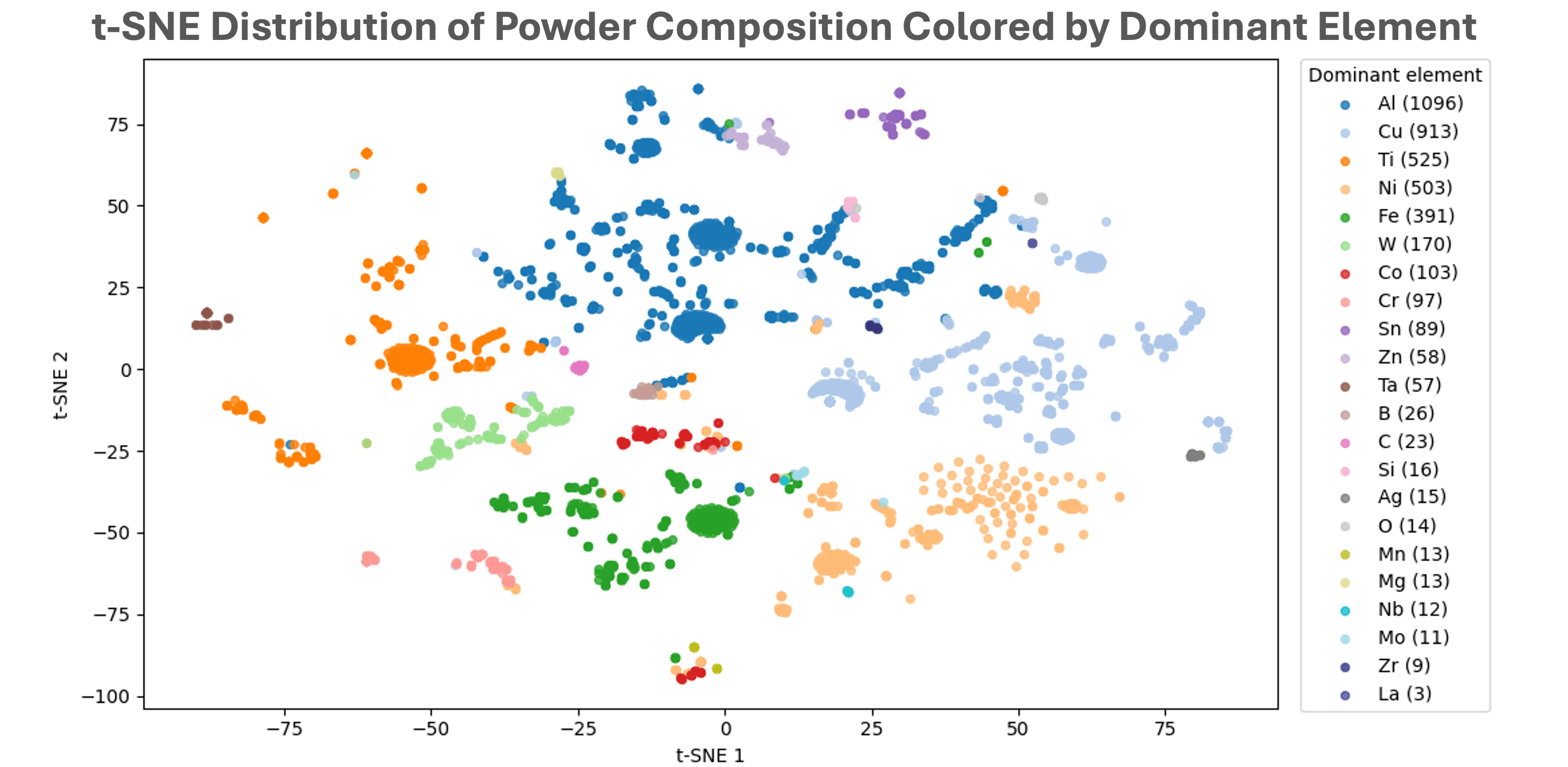}
    \caption{}
    \label{fig:tsne}
  \end{subfigure}
  \caption{(A) Visualization of extracted values per mechanical property, where red indicates experiments that report uncertainty (e.g., error bars or $\pm$ values) and gray indicates experiments reported without uncertainty. (B) t-SNE visualization of the distribution of powder-blend compositions extracted from the curated cold-spray literature dataset, colored by the dominant element.}
  \label{fig:uncertaintyAndDistribution}
\end{figure}

\noindent\textbf{Uncertainty Aware.} In addition to extracting individual values, H\textbf{U}GO is designed to also capture the experimental uncertainty reported alongside those values. When available, HUGO-CS includes uncertainty as reported by the source, including $\pm$ values, error bars, standard deviations, standard errors, and confidence intervals. As shown in Fig.~\ref{fig:uncertainty}, the frequency of uncertainty reporting varied significantly by target variable. For example, nano-hardness was reported most often in the literature with its uncertainty, with 62.7\% of values including uncertainty, whereas deposition efficiency (DE) is reported with uncertainty least often, with only 16.2\% of values including reported uncertainty. While not all measurements include uncertainty, preserving it when available enables downstream users to separate value estimates from aggregated results and to incorporate experimentally reported variability into meta-analysis, uncertainty quantification, and robust model training.

\noindent\textbf{General Purpose.} Compared to prior datasets that are often narrowly focused on a specific task or sub-domain, such as Ni-based superalloys \cite{sudigdo2025cold}, HU\textbf{G}O-CS  is designed to be a general-purpose dataset spanning diverse materials and processing conditions covering Cold Spraying (CS). For example, HUGO-CS includes numerous majority powder primary elements (e.g., Al, Cu, Ti). Further, within a given primary element, HUGO-CS captures many distinct alloy families and compositions rather than collapsing them into a single categorical label (e.g., Pure Al, Al~6061, Al~7075), as shown in Fig.~\ref{fig:tsne}, displaying a t-SNE projection of extracted powder compositions colored by the dominant elemental component.

\noindent\textbf{Observational.} As an \textit{observational} dataset, HUG\textbf{O}-CS captures reported cold-spray mechanical properties in a structured format, along with the processing provenance needed to reproduce, compare, or model these results. The extraction schema was developed iteratively with domain input, combining knowledge from domain-experts  with repeated refinement and re-labeling to ensure that experimental factors observed across the literature were represented in the final feature set. Wherever possible, these factors are recorded as atomic fields rather than merged free-text, enabling consistent sorting, filtering, and subset selection across studies. Finally, each experimental record includes metadata that ties it back to its source article, including DOI information (or an alternative identifier when necessary). This allows users of HUGO-CS to trace measurements to the original publication for broader context and detailed descriptions of the study.

\subsection{Targeted Manual Relabeling with Hierarchical Risk Mitigation}
In this work, the Hierarchical Risk Mitigation (HRM) module, described in Sec.~\ref{sec:HRM_Main}, was used to concentrate manual labeling on the articles whose LLM extractions were most likely to contain errors. As shown in Table~\ref{tab:curation_summary}, this was performed in multiple stages, identifying extractions with incorrect syntax, with an incorrect schema, with outlier values, and those that likely missed experiments presented only in figures, and not text. Flagged articles were then manually labeled to replace the erroneous extractions with corrected hand-labels. Table~\ref{tab:curation_summary} reports these changes at both the article and experiment levels. At the article level, \textit{added}, \textit{re-labeled}, and \textit{deleted} indicate (i) sources that did not previously have a usable LLM extraction but were found by manual labeling to contain at least one eligible experiment, (ii) sources where both the LLM extraction and hand-labeled extraction contained at least one eligible experiment, and (iii) sources where the LLM extraction contained at least one experiment, but manual re-labeling identified no eligible experiments. At the experiment level, these terms indicate (i) newly added experiments, (ii) existing experiments that were modified or confirmed through hand-labeling, and (iii) LLM-extracted experiments that were removed during hand-labeling.

\begin{table*}[!h]
\centering
\small
\setlength{\tabcolsep}{6pt}
\renewcommand{\arraystretch}{1.15}
\caption{Overview of hybrid-labeling process per step, highlighting the results of LLM and hand-labeled as a result of automated flagging from incorrect output structures, high-chance of data in figures, missing features, statistical outliers, and other hand-labeled articles.}
\label{tab:curation_summary}
\begin{adjustbox}{max width=\textwidth}
\begin{tabular}{l|c|ccc|ccc}
\toprule
& & \multicolumn{3}{c|}{\textbf{Articles}} & \multicolumn{3}{c}{\textbf{Experiments}} \\
\cmidrule(lr){3-5}\cmidrule(lr){6-8}
\textbf{Pipeline Step} & \textbf{Process Type} & \textbf{Added} & \textbf{Re-labeled} & \textbf{Deleted} & \textbf{Added} & \textbf{Re-labeled} & \textbf{Removed} \\
\midrule
Initial LLM Labeling & LLM & 1074 & 0 & 0 & 3366 & 0 & 0 \\
- Syntax Flagging Correction & Algorithmic / Manual & +37 & 0 & 0 & +261 & 0 & 0 \\
- Completeness Flagging Correction & Algorithmic / Manual & 0 & 32 & -8 & +122 & 205 & -68 \\
- Outlier Flagging Correction & Algorithmic / Manual & 0 & 32 & 0 & +54 & 128 & -13 \\
- Figure Flagging Correction & Algorithmic / Manual & +32 & 33 & -4 & +554 & 97 & -9 \\
- Other Flagging Correction & Manual & +3 & 37 & -5 & +146 & 118 & -29 \\
- Held-Out Validation & Manual & 0 & 19 & -1 & +11 & 69 & -8 \\
- Empty Result Removal & Algorithmic & 0 & 3 & 0 & 0 & 0 & -4 \\
- Key Realignment & Algorithmic & 0 & 36 & 0 & 0 & 78 & 0 \\
- Categorical String Mapping & Manual / Algorithmic / LLM & 0 & 1124 & 0 & 0 & 4383 & 0 \\
- Continuous Composition Mapping & Manual / Algorithmic / LLM & 0 & 660 & 0 & 0 & 2425 & 0 \\
- Known Composition Imputation & Manual / Algorithmic / LLM & 0 & 583 & 0 & 0 & 2202 & 0 \\
- Multi-Powder Composition Blending & Manual / Algorithmic / LLM & 0 & 1070 & 0 & 0 & 4157 & 0 \\
\midrule
\textbf{LLM-Labeled Dataset Size} & --- & 1074 & --- & --- & 3366 & --- & --- \\
\textbf{Hand-Labeled Dataset Size} & --- & +68 & 157 & -18 & +1148 & 617 & -127 \\
\midrule
\textbf{Hybrid-Labeled Dataset Size} & --- & 1124 & --- & --- & 4383 & --- & --- \\
\bottomrule
\end{tabular}
\end{adjustbox}
\end{table*}

\subsubsection{Syntax Errors: Correcting Unparsable Outputs}
The first of these stages was structurally invalid JSON outputs, containing malformed or truncated responses that could not be reliably parsed into an experimental record. All articles with a structurally invalid JSON output were flagged for manual labeling (Sec.~\ref{sec:HRM_Stage1}). As shown in Table~\ref{tab:curation_summary}, this syntax-correction step identified 37 articles whose LLM outputs were unreadable and required manual labeling. Of these, 33 articles contained experimental values, adding +261 experiments from manual labeling, while 4 articles were found to contain no novel experimental results.

\subsubsection{Completeness Errors: Correcting Non-Compliant Schemas}\label{sec:ResultsOutliers}
After structurally invalid experiments, any experiment containing an incorrect schema with missing attributes or unexpected attributes was re-labeled. In some  cases, these errors could be corrected without manual intervention by realigning mis-keyed fields to the expected schema, reducing the number of outputs routed to hand-labeling (Sec.~\ref{sec:HRM_Stage2}). Namely, as shown in Table~\ref{tab:curation_summary}, 36 articles (78 experiments) were corrected by string-similarity key realignment, reducing the burden of manual labeling. After realignment, 40 articles were still flagged, requiring re-labeling. Of these, 8 were found to contain no novel experimental results, removing 68 experiments, while the remaining 32 hand-labeled articles resulted in 205 experiments re-labeled and +122 new experiments added.

\subsubsection{Outlier Errors: Correcting Anomalous Experiments}
After re-labeling syntax and completeness errors, correcting structure-based extraction failures, articles were flagged and re-labeled for content-based errors using statistical outliers. As shown in Table~\ref{tab:curation_summary}, this outlier-correction step identified 32 articles for manual labeling, resulting in +54 added experiments, 128 re-labeled experiments, and 13 removed experiments. Outside of these 32 articles, additional sources were flagged but, ultimately, not re-labeled. For example, Wang et al.~\cite{wang2020high} reported both cold spray and cold spray-friction stir processing (CS-FSP) experiments for pure aluminum. The LLM extracted an elongation of 60.3\% for one experiment, which was flagged as an outlier relative to other pure-aluminum samples (class average: 4.7\%). A comparison between the LLM extraction and the source article confirmed that (i) 60.3\% was correctly extracted, (ii) the anomalous value could be explained by CS-FSP processing, and (iii) the necessary FSP-related parameters were captured in the same experimental record. As a result, this article was not re-labeled. In cases like these, the article was inspected, but not fully hand-labeled. Thus, the corresponding experiments remain marked as LLM-labeled in HUGO-CS, since the manual review was limited to the flagged outlier feature. 

% In some cases, identified outliers appeared to reflect errors in the original source reporting. For example, Cizek et al.~\cite{cizek2016potential} reported Young's modulus values ranging from 4,641 $\pm$ 57 to 8,008 $\pm$ 285~GPa. Given that the next highest reported modulus in the dataset was 452~GPa (TiC modulus determined by nanoindentation)~\cite{georgiou2017wear}, the 8,008~GPa value was likely a unit or decimal error (e.g., 8.008~GPa or 8,008~MPa). Correcting or removing these values was considered, but ultimately they were left as-is, since they were correctly extracted from the source and HUGO-CS is intended to remain observational.

\subsubsection{Coverage Errors: Correcting Results Only Reported in Figures}
After errors in existing extractions were corrected, HRM-flagged articles likely to contain missing experiments, because they only reported in figures, rather than text. Using a \textit{wEV} threshold of 2.5, a total of 69 articles were flagged for manual labeling, as shown in Table~\ref{tab:curation_summary}. This included 32 documents where the initial LLM extracted no experimental values, and 33 documents that contained at least one LLM-extracted experiment but that were expanded after the manual labeling. Overall, this stage added +554 previously missed experiments and re-labeled an additional 97 experiments, improving coverage beyond text-only extraction. Additionally, 9 articles had LLM-labeled experiments that were pruned during manual review as survey papers rather than novel experimental results.

\subsubsection{Unflagged Errors: Additional Manual Labeling}
In addition to the articles flagged by HRM, a small set of unflagged articles were also manually labeled. These included documents used during initial prompt optimization as well as articles used for training manual labelers. Several articles that required manual correction were discovered during data cleaning steps (Sec.~\ref{sec:postProcessing}). For example, in Pandey et al.~\cite{pandey2024hvof}, the authors report experimental results for HVOF, a high-velocity powder deposition process that heats the powder to much higher temperatures prior to deposition and produces substantially different mechanical properties. The LLM incorrectly interpreted these results as cold spraying and extracted them, labeling the spray system as ``HVOF''. Manual labeling determined that no novel cold-spray experiments were present in that entry, causing it to be removed from the dataset. As shown in Table~\ref{tab:curation_summary}, these additional hand-labels included 45 source articles, resulting in +146 added experiments, 118 re-labeled experiments, and -29 removed experiments.

\subsection{LLM Extraction Performance Evaluation}
For an unbiased performance evaluation of LLM-based extraction on the remaining articles that were not flagged by HRM, an additional 20 articles were randomly sampled and hand-labeled. Of these, 19 contained experimental results that met the inclusion/exclusion criteria, yielding 80 ground-truth experiments for direct comparison against the corresponding LLM extractions. On this held-out testing set, the LLM extracted 77 experiments. Of these, 69 matched a ground-truth record, yielding a precision of 89.61\% and a recall of 86.25\%. Analysis of the false-positive experiments, where the LLM extracted an experiment that could not be matched to a ground-truth record, revealed that many failures were lack of following the instruction rather than hallucinated values. For example, in Diaz et al. \cite{diaz2025graphene}, the authors performed six cold-spray experiments and reported both tensile properties and microhardness for each condition. However, the LLM failed to group these results by shared processing parameters and instead reported twelve ``experiments,'' splitting tensile values and hardness values into separate records. While this reduces the ability to link material properties across target variables, it reflects a failure to correctly group experimental values rather than an incorrect extraction of the underlying measurements. Cross-referencing with the article and the hand-labeled output confirmed that these additional records contained correctly extracted values. Alternatively, in Kumar et al. \cite{kumar2022study}, the LLM extracted two Fe-based UTS experiments (64.58 $\pm$ 42.07~MPa and 109.42 $\pm$ 50.56~MPa), but these values summarize experiments reported in Xie et al.~\cite{xie2022improvement} and, therefore,
they are not novel to the source article. This violates the inclusion criteria and inflates the count of the extracted experiments. In both cases, the extracted numbers correspond to real reported cold-spray measurements, while the errors arose from how those measurements are grouped or filtered rather than from hallucinated values.

% Note to self: Diaz et al. is Article 875
% Note to self: Kumar et al. is Article 286
% Note to self: Bai et al. is Article 2191

With respect to recall, many of the missed experiments occurred because their experimental results were only presented in figures and unavailable to the text-extraction pipeline. For example, in Bai et al. \cite{bai2018microstructure}, the authors tested four samples to quantify how nickel content impacted microhardness, and only reported these values in a figure. HUGO correctly identified that values were likely missing. However, since only microhardness values were missing, with no other features tested, and microhardness was the highest frequency feature, this article scored a \textit{wEV} of 0.093, which did not exceed the $T=2.5$ threshold to trigger manual labeling. With a lower $T$ value, this would have been flagged, re-labeled by hand, and
would have contained no missing experiments.

At a feature level, extracted experiments averaged an accuracy of 94.55\% across schema fields. Accounting for sparsity and evaluating only non-empty fields, these extractions averaged a precision of 84.95\% and a recall of 79.07\%. A closer evaluation revealed two primary failure cases. First, as discussed above, the model was unable to extract information reported exclusively in figures. However, the figure-flagging stage used for manual labeling was primarily designed to recover missing \textit{target variables} (e.g., hardness, yield strength, porosity).

While effective for these outcomes, it did not fully address the same issue for \textit{experimental parameters}. In particular, tensile specimen geometry is often provided only as an engineering diagram rather than explicitly stated in text, resulting in substantially lower extraction accuracies of 60.87\% for gauge width, 73.91\% for gauge length, and 65.22\% for gauge thickness. Second, a frequent source of error was excessive verbosity from the LLM compared to human labelers. During schema construction, fields such as \textit{Powder Morphology Other} and \textit{Additional Manufacturing Process Description} were included to capture relevant factors that do not fit existing attributes. In practice, the LLM frequently populated these fields with summaries of information already captured elsewhere in the schema, whereas human labelers more often left these fields empty when no additional distinct information was present, resulting in a lower agreement with the hand-labeled extractions.

\subsection{Data Cleaning and Unification} \label{sec:ResultsPostProcessing}
During LLM-based and manual extraction, features were recorded in their as-reported form from the source paper. As result, many fields contain free-text with inconsistent syntax, naming conventions, and different levels of detail and metrics across sources, as further discussed in Sec.~\ref{sec:postProcessing}. If left unprocessed, downstream tasks such as literature search, analysis, or modeling would require additional parsing steps (e.g., regular expressions or a secondary LLM) to identify and process semantic similarities. To address this issue, the PIR modules were applied to key features by mapping free-text to standardized categorical labels and normalizing numeric values into consistent continuous representations, as further described below.

For {\it categorical variables,} data cleaning procedures were deployed to standardize free-text into consistent labels for the majority, secondary, and tertiary feedstocks (material type and primary element), as well as powder production method, process gas, tensile orientation, cold-spray system, and post-deposition treatment type. As shown in Table~\ref{tab:mapping_reduction}, this consolidation resulted in a significant reduction for some fields, such as 91.0\% for the spray system used, or 80.4\% for the powder production method. For many of these attributes, they were directly replaced by the standardized representation, such as ``Al-6061 powder'' to ``Al 6061.'' 

However, in the case of post-deposition treatment, an additional feature was added rather than replaced. Namely, when the deposition treatment was extracted, fields were included for the temperature, duration, and cool-down time of up to three treatment cycles, as well as a description field to describe the process. This description contained the treatment type (e.g., annealing, sintering, solutionizing) and more detailed process information that could not directly fit the temperature, duration, nor cool-down time. A direct mapping to a standardized categorical value would lose this additional information. As a result, HUGO-CS retains the original \textit{Deposit Post-Treatment Description} as well as the newly created \textit{Treatment Categorical}.

\begin{table*}[!h]
\centering
\small
\setlength{\tabcolsep}{6pt}
\renewcommand{\arraystretch}{1.15}
\caption{Quantitative overview of consolidated features, highlighting the consolidation of text-based semantic information to concise categorical values.}
\label{tab:mapping_reduction}
\begin{adjustbox}{max width=\textwidth}
\begin{tabular}{l|c|c|c}
\toprule
\textbf{Feature} & \textbf{Unique Values (Raw)} & \textbf{Unique Values (After Consolidation)} & \textbf{Reduction} \\
\midrule
Majority Powder Material & 680 & 211 & 69.0\% \\
Majority Powder Primary Element & 50 & 40 & 20.0\% \\
Secondary Powder Material & 251 & 106 & 57.8\% \\
Secondary Powder Primary Element & 36 & 33 & 8.3\% \\
Tertiary Powder Material & 40 & 33 & 17.5\% \\
Tertiary Powder Primary Element & 18 & 18 & 0.0\% \\
Powder Production Method & 306 & 18 & 94.1\% \\
Process Gas & 56 & 11 & 80.4\% \\
Tensile Test Orientation & 117 & 14 & 88.0\% \\
Cold Spray System & 635 & 57 & 91.0\% \\
Post-Deposition Treatment & 679 & 7 & 99.0\% \\
\bottomrule
\end{tabular}
\end{adjustbox}
\end{table*}

For continuous values, including the majority, secondary, and tertiary feedstock elemental compositions, these were mapped to structured numerical attributes to preserve detailed chemistry information and enable composition-level comparisons. However, to ensure these chemistries could be consistently compared across sources, this mapping required a multi-stage workflow, as shown in Fig.~\ref{fig:CompositionWorkflow}. 
In total, 2,955 composition strings were converted to a 50-element structured representation. Of these, 2,805 were already reported in wt.\%, while 150 had to be converted to wt.\% using known element-specific constants. Additionally, 1,603 experiments used blended feedstocks, with 1143 containing two feedstocks and 460 containing three feedstocks.

\subsection{Secondary Labeling Process for Improved Experimental Provenance}\label{sec:ResultsYS_Relabel}
Across published cold-spray literature, different authors and labs use a variety of experimental techniques to measure deposited material properties. For example, yield strength can be measured directly from uniaxial tensile testing or indirectly estimated using indentation-based approaches. These values are often reported under the same label despite reflecting different measurement procedures. However, these different techniques can produce different measured values. To improve experimental provenance, explain numerical outliers, and support uncertainty estimates, each experiment was processed by an additional LLM prompt to identify and flag results likely derived from an alternate process, and are stored in the HUGO-CS dataset. This secondary labeling flagged 156 modulus values, 38 yield strength values, 19 microhardness values, and 18 ultimate tensile strength values, allowing downstream users to keep, remove, or assign larger uncertainties to these values.

\section{Discussion}
This work introduces HUGO, a hybrid-labeling framework for extracting experimental values from the literature. To balance the efficiency of LLM-based labeling with the accuracy of manual extraction, HUGO leverages a Hierarchical Risk Mitigation (HRM) strategy to detect and flag articles likely to contain errors. This allows manual effort to be concentrated where it is expected to have the most positive impact. Using the HUGO framework, HUGO-CS was produced as a large-scale machine-readable dataset for cold spraying, in addition to an analysis of the  labeling performance and post-processing steps required to produce a machine-readable dataset.

\subsection{Open Accessibility of HUGO for Reproducibility and Future Expansion}
All code necessary to extend HUGO's hybrid-labeling process for future cold-spraying works, or adapt it to new domains, has been released with the accompanying GitHub repository. This includes code for article extraction, automated hierarchical document flagging of high-risk documents, assisted schema generation, and assisted string mapping. As a result, practitioners could reproduce the curation process used here or extend HUGO-CS to include newly published cold-spray studies that were not available at the time of this release. Similarly, HUGO could also be applied to convert newly published literature into structured machine-readable experimental records in other application domains, expanding the utility of existing publications for data-driven meta-review, modeling, and process-optimization workflows.

\subsection{Potential Applications of HUGO-CS to Cold Spraying}
\begin{figure}[!h]
  \centering
  \begin{subfigure}[t]{0.49\textwidth}
    \centering
    \includegraphics[width=\linewidth]{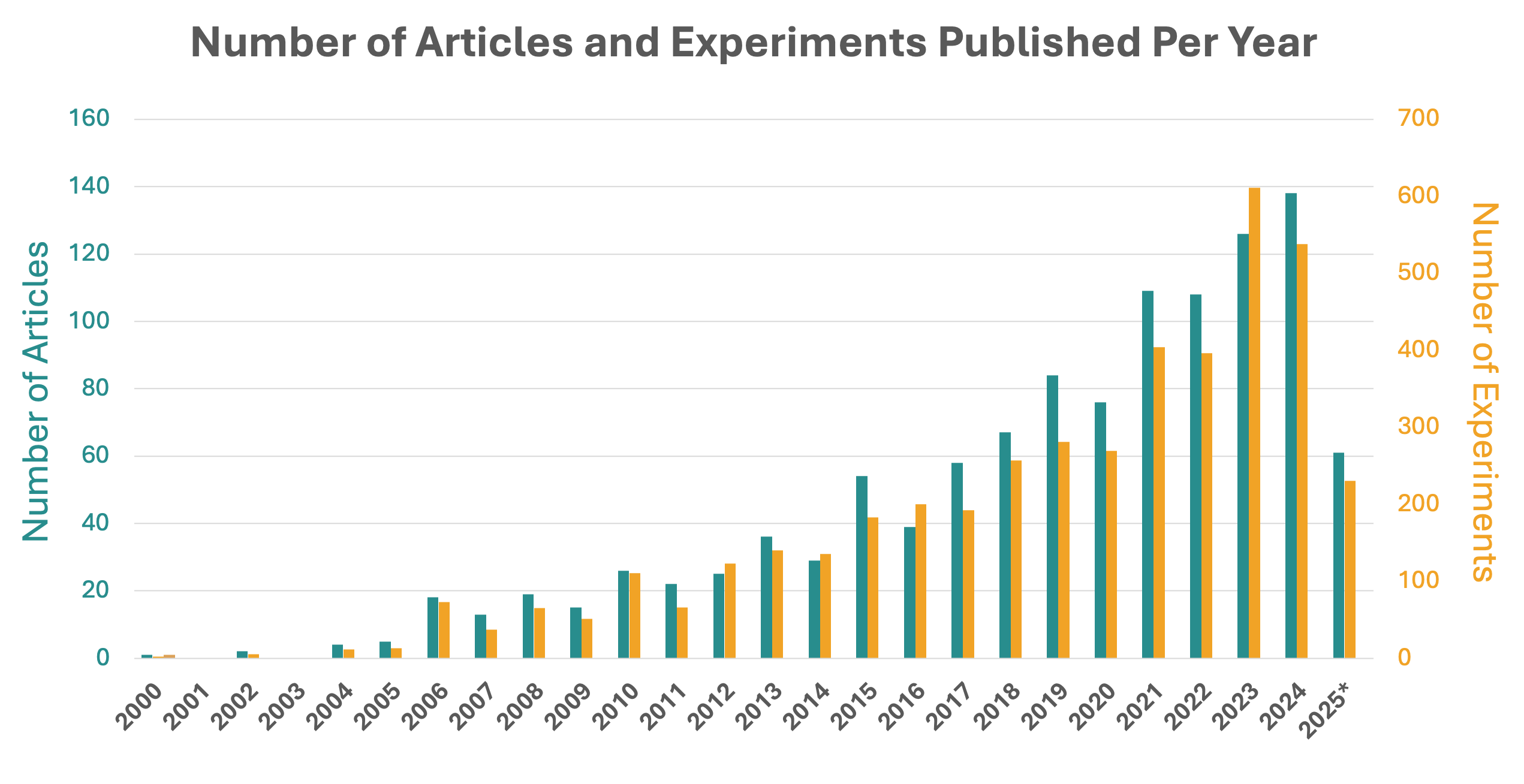}
    \caption{}
    \label{fig:publicationsPerYear}
  \end{subfigure}\hfill
  \begin{subfigure}[t]{0.49\textwidth}
    \centering
    \includegraphics[width=\linewidth]{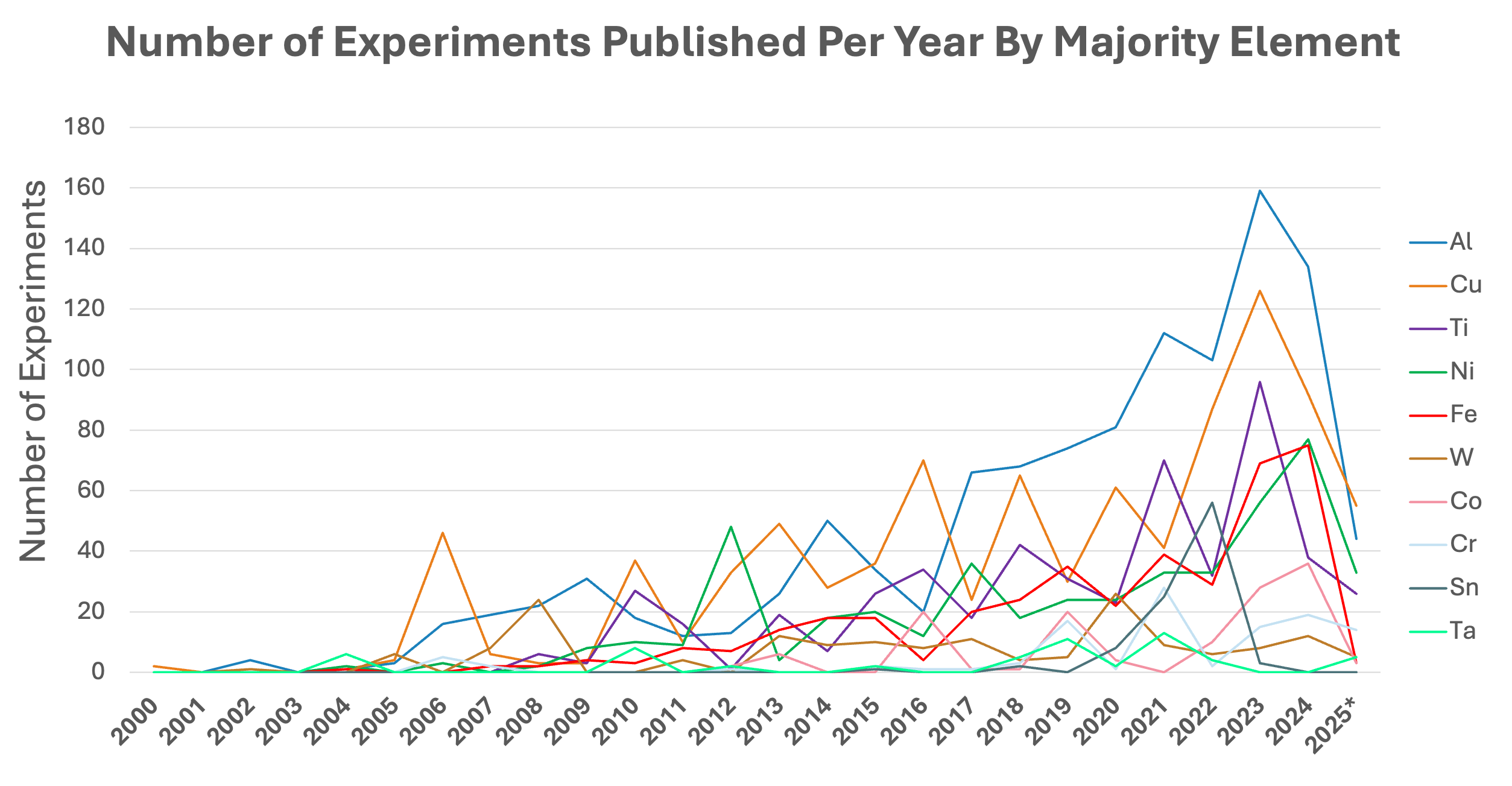}
    \caption{}
    \label{fig:publicationsPerYearAlloy}
  \end{subfigure}

  \caption{(a) Number of cold spray articles and extracted experiments published per year in HUGO-CS, and (b) Number of extracted experiments published per year, stratified by majority powder primary element.}
  \label{fig:timeBasedAnalysis}
\end{figure}

\begin{figure}[!h]
  \centering
  \begin{subfigure}[t]{0.49\textwidth}
    \centering
    \includegraphics[width=\linewidth]{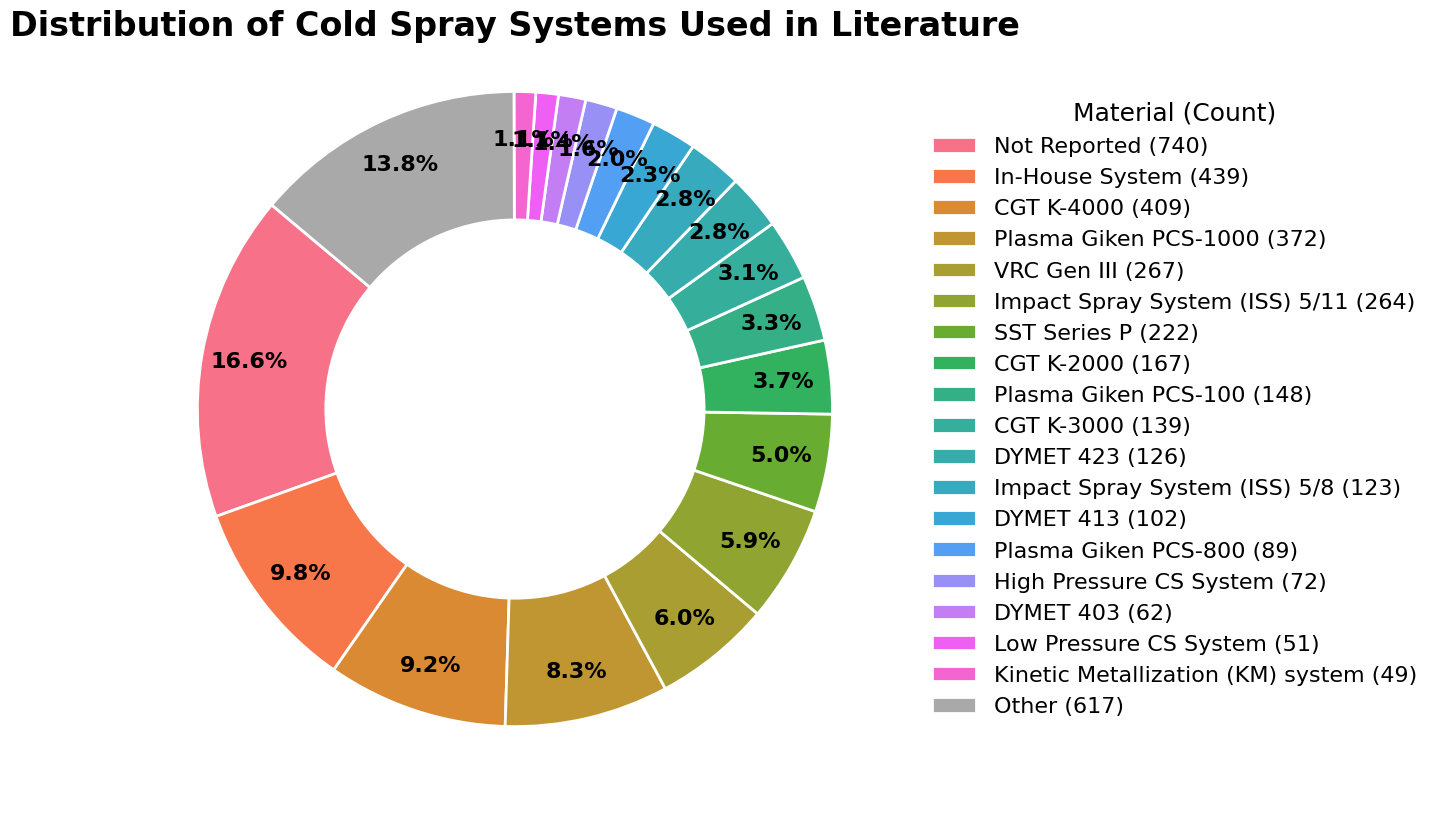}
    \caption{}
    \label{fig:suba_spraysystem}
  \end{subfigure}\hfill
  \begin{subfigure}[t]{0.47\textwidth}
    \centering
    \includegraphics[width=\linewidth]{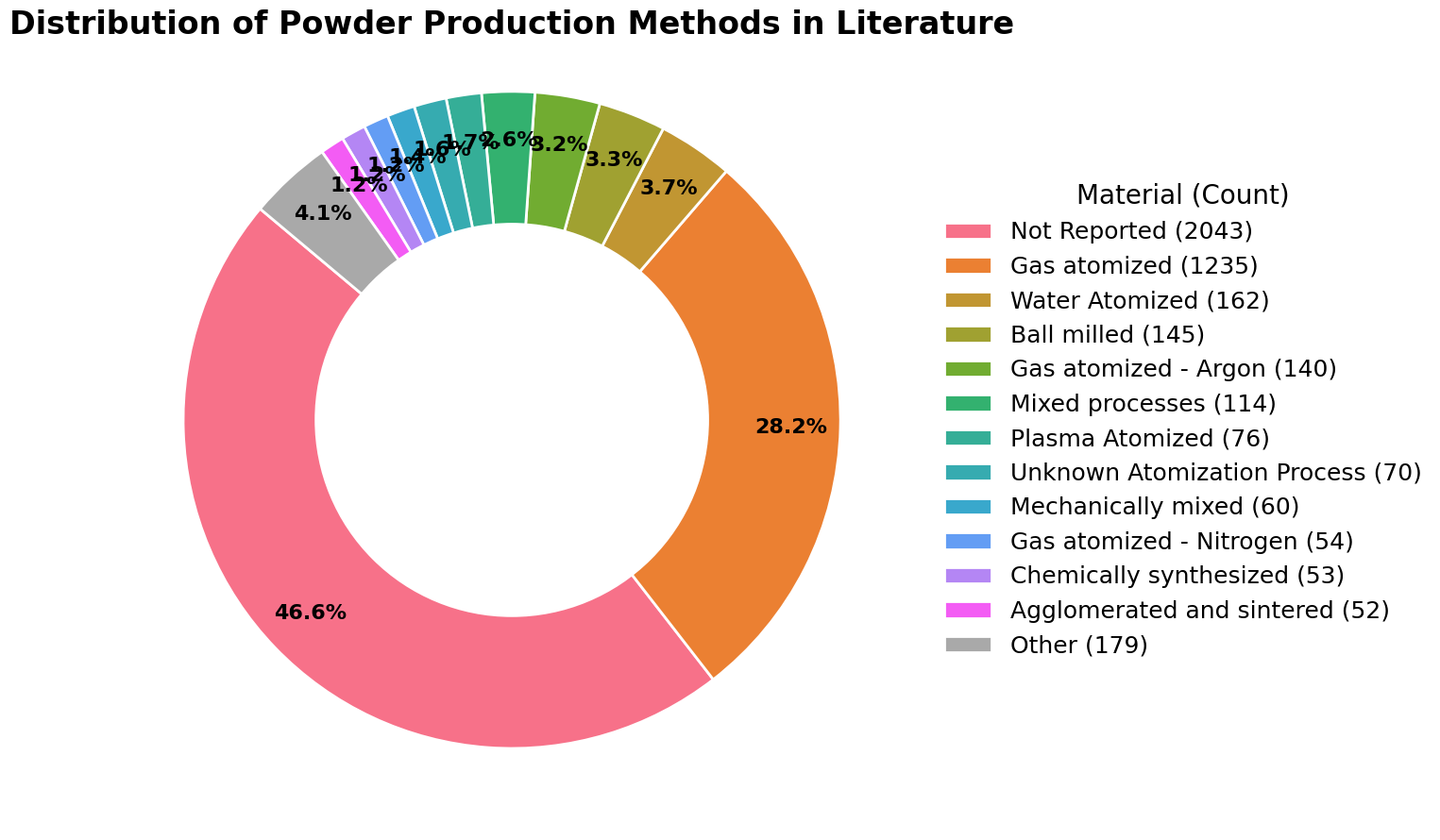}
    \caption{}
    \label{fig:subB_productionDistribution}
  \end{subfigure}

  \caption{(A) Visualization of the distribution of cold-spray systems reported across HUGO-CS, and (B) distribution of powder production methods reported across HUGO-CS.}
  \label{fig:distributionDemonstration}
\end{figure}

The creation of HUGO-CS represents a significant increase in the size and the level of detail accessible for cold-spray results, offering numerous opportunities to the cold-spraying community. For example, as shown in Figure \ref{fig:timeBasedAnalysis}, meta-reviews can be conducted to trace the increase in experimental work over time (Figure \ref{fig:publicationsPerYear}), or 
even further refined 
%further isolated to
to track the growth over time of specific types of cold spraying (Figure \ref{fig:publicationsPerYearAlloy}). Alternatively, users can analyze the distribution of processing parameter design decisions, such as the frequency of each cold spray system reported across studies (Fig.~\ref{fig:suba_spraysystem}) or the powder production methods used to generate feedstock powders (Fig.~\ref{fig:subB_productionDistribution}).

% \subsection{Distribution of Reported Values}
% As an observational dataset, this work contains the biases and imbalances of published literature. For example, as shown in Fig.\ref{fig:subA_materails}, almost 60\% of samples are Al-, Cu, or Ti-based alloys. Similarly, atomized is X\%, with X as gas atomiszes, X as Gas atomized, specified iwth argon, X as gas-atomized as specified with nitrogen, and X as water atomized, and X with plasma-atomized. Similarly, aproximatley 75\% of samples were recorded with no post-deposition treatment. 

\subsubsection{Modeling Applications}
In addition to retrospective analysis, HUGO-CS can be used for modeling and process optimization. To demonstrate this, two proof-of-concept models were trained: a multi-material microhardness model and a targeted yield strength model for aluminum alloys. Here, the microhardness model was trained on 2431 experiments with reported HV hardness values. In comparison, the yield strength model was trained on 58 experiments with reported yield strength values whose feedstock contained at least 85\% aluminum. In both cases, model evaluation was separated at the article level, preventing data leakage from experiments reported in the same source across validation folds. For microhardness, CatBoost was trained using grouped 5-fold cross-validation at the article level after excluding experiments involving additional post-processing techniques or non-standard test methods. Scikit-learn feature selection was then applied within training folds, dropping the 50\% least important features before retraining the model. For yield strength, due to the limited training size, leave-one-article-out-cross-validation (LOAO-CV) was performed, where a separate model was trained for each held-out article and predictions were aggregated across folds. Using LOAO, a Gradient Boosting regressor was trained using k-nearest neighbor imputation for missing features, as well as feature engineering to encode the maximum nominal yield strengths of the five nearest ASM alloys based on composition-space similarity.

\begin{figure}[!h]
  \centering
  \begin{subfigure}[t]{0.47\textwidth}
    \centering
    \includegraphics[width=\linewidth]{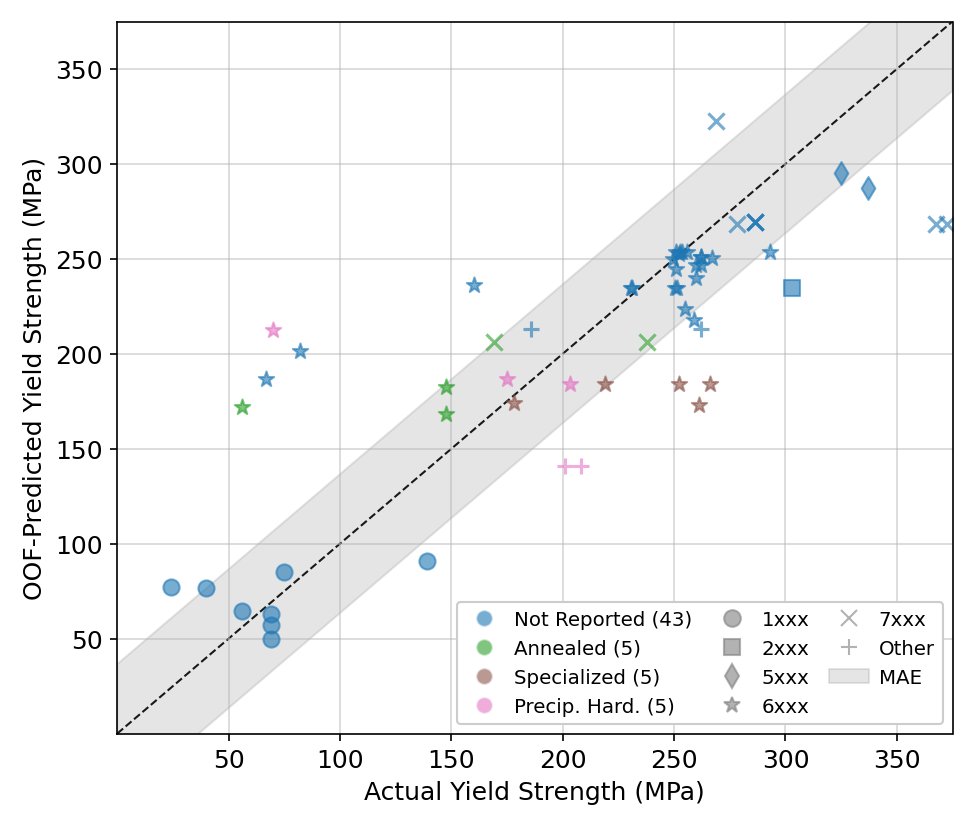}
    \caption{}
    \label{fig:al_yieldstrength_graph}
  \end{subfigure}\hfill
  \begin{subfigure}[t]{0.47\textwidth}
    \centering
    \includegraphics[width=\linewidth]{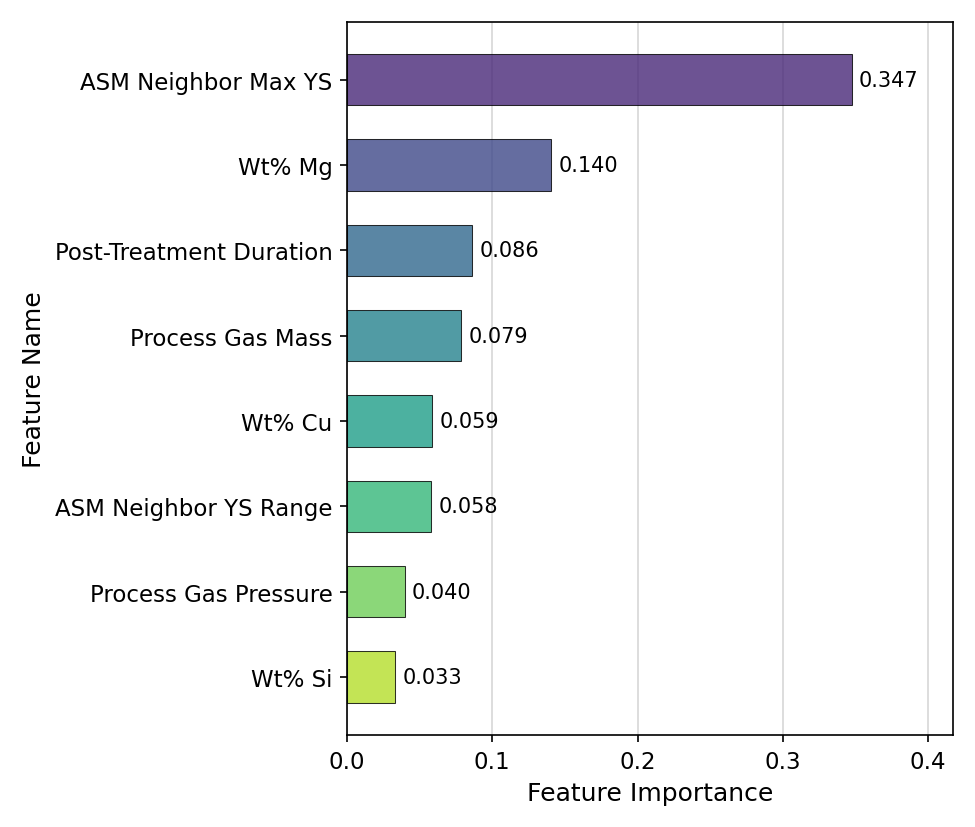}
    \caption{}
    \label{fig:al_yieldstrength_feature}
  \end{subfigure}

  \caption{(A) Visualization of leave-one-article-out (LOAO) yield strength predictions for experiments with over 85\% aluminum, and (B) highest ranked feature importances for the Gradient Boosting regressor used to predict aluminum yield strength.}
  \label{fig:al_yield}
\end{figure}

Averaged across LOAO-CV folds, the aluminum yield strength model scored a mean absolute error (MAE) of 36.6  MPa and an $R^2$ of 0.66. As shown in Fig.~\ref{fig:al_yieldstrength_graph}, the largest errors occurred for deposits with specialized treatments, such as precipitation-hardened deposits, where training coverage was sparser. In contrast, predictions were more accurate for the lower-strength 1xxx-series alloys and the higher-strength 6xxx-series alloys, which were predominantly Al~6061. Analyzing individual feature importance, as shown in Fig.~\ref{fig:al_yieldstrength_feature}, the weighted nominal maximum yield strength of the nearest ASM alloys was the most important predictor, enabling delta-learning on top of known bulk alloy behavior instead of learning from scratch. In addition, Mg percentage, treatment time, and carrier gas were quite important to inference predictions. 

\begin{figure}[!h]
  \centering
  \begin{subfigure}[t]{0.47\textwidth}
    \centering
    \includegraphics[width=\linewidth]{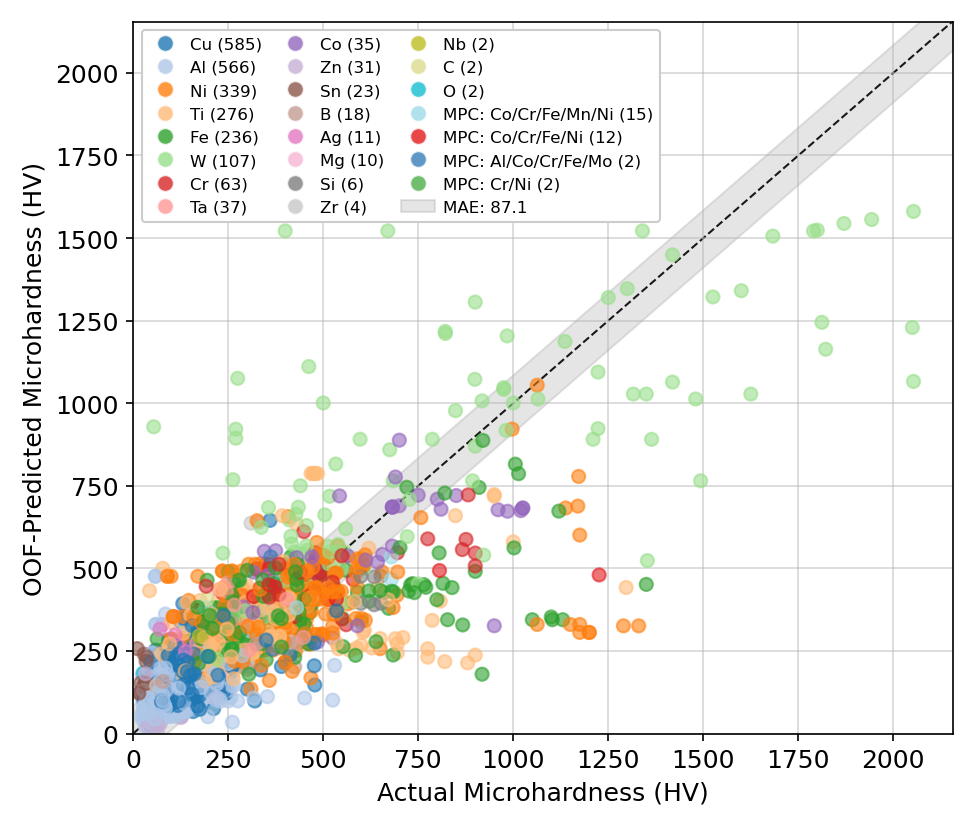}
    \caption{}
    \label{fig:microhardness_graph}
  \end{subfigure}\hfill
  \begin{subfigure}[t]{0.47\textwidth}
    \centering
    \includegraphics[width=\linewidth]{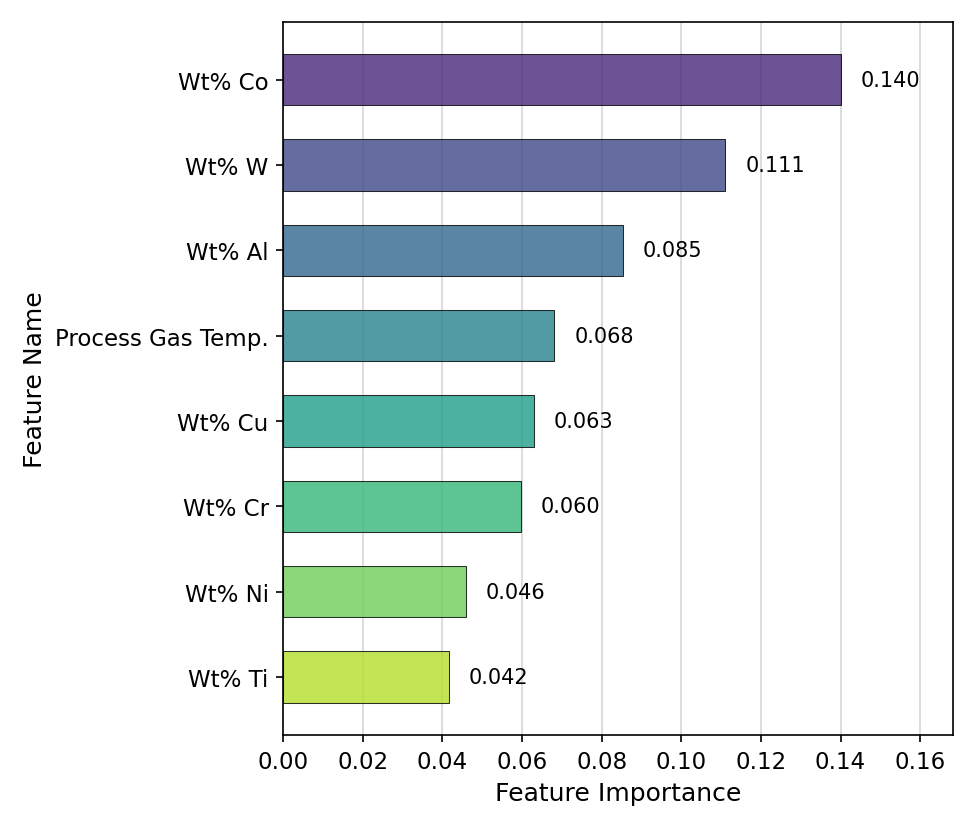}
    \caption{}
    \label{fig:microhardness_feature}
  \end{subfigure}

  \caption{(A) Visualization of predicted versus actual microhardness values for the multi-class hardness model, and (B) highest ranked feature importances for the CatBoost regressor used to predict microhardness.}
  \label{fig:microhardness}
\end{figure}

For the multi-material microhardness predictions, as shown in Fig.~\ref{fig:microhardness_graph}, the CatBoost model scored an $R^2$ of 0.65, with an MAE of approximately 87.12 HV. This highlights a general ability to predict hardness trends, but predictions were still subject to noise. Tungsten-based alloys, including Tungsten-Carbide (WC) and Tungsten-Carbide Cobalt (WC-CO), had some of the largest errors, likely due to the significantly larger chemical diversity and broader hardness ranges compared to the rest of the samples. Analyzing feature importance, as shown in Fig.~\ref{fig:microhardness_feature}, composition-driven features were the dominant predictors, with seven of the top eight features relating to composition, including tungsten and cobalt weight fractions ranking as the two highest. This indicates the model primarily learned composition-based trends, as well as some experimental parameters, such as process gas temperature, matching the targeted aluminum yield strength results. Additionally, the prevalence of these composition-based features as dominant predictors highlights the value of extracting and post-processing feedstock composition, which many prior works excluded, as shown in Table~\ref{tab:sota_comparison}.

\subsection{Traceability and Release Scope}
HUGO-CS is a machine-readable, structured representation of experimental cold-spray values reported in literature, with each record linked back to its primary source (as discussed in Sec.~\ref{sec:metadata_traceability}). The GitHub repository contains the complete HUGO software including the extraction pipeline and the extracted HUGO-CS dataset. However, to avoid copyright violations, the HUGO-CS release, while containing the extracted experimental records, does not include full-text papers, figures, or formatted tables from the original publications. Additionally, to avoid copying content from other sources, this work took the precaution to rephrase any lengthy entries. That is, for extracted free-text fields, entries exceeding 30 characters were manually inspected. In cases when they substantially overlapped with the source material or contained non-experimental prose, they were paraphrased and compressed to retain only experimentally relevant content while avoiding reproduction of text from their source publications.

\subsection{Limitations}
When developing a structured schema for a data-driven representation, one of the key difficulties is capturing heterogeneous experimental procedures within a single standardized form. Cold spray studies frequently include additional processing steps that substantially alter deposit properties (e.g., powder conditioning, feedstock blending, post-deposition treatments, and hybrid processes such as laser assistance, micro-forging, shot-peening, HIP, hot rolling, or friction-stir processing), as well as variable testing protocols. While the HUGO-CS schema was designed to capture many of these factors, it does not guarantee that every experiment is fully represented under the current feature set. As discussed in Sec.~\ref{sec:YS_Relabel}, reported values like yield strength or elastic modulus values can depend strongly on the testing procedure and may be derived from alternative measurement methods. HUGO-CS partially mitigates this through a secondary labeling stage, but reflects the broader diversity in experimental design and reporting conventions. 

Additional alignment gaps are present where experiments exceed the capacity of the specified schema. For example, to capture blended feedstocks, HUGO-CS records the primary, secondary, and tertiary feedstock compositions, which is sufficient in nearly all cases. However, in Yao et al. \cite{yao2023microstructure}, the authors mechanically mixed six elemental powder feedstocks, a special case that would not be fully represented by our three-tiered modeling. Thus, to preserve the reported chemistry despite this issue, the two dominant components were recorded as the primary and secondary feedstocks, while the remaining four lower-fraction components were recorded in the tertiary field as a consolidated multi-powder entry. Similarly, Poza et al. \cite{poza2014mechanical} performed laser remelting as a post-deposition step, which did not cleanly map to existing LA-CS or thermal treatment fields and was therefore stored in the \textit{Additional Manufacturing Processes} field as a free-text description. While HUGO-CS was designed to be as representative as possible, edge cases like these still can exist, requiring less precise schema adherence to preserve the reported experimental information.

%Note to self: Yao et al. is Article 1829
%Note to self: Poza et al. is Article 1366

Another limitation is that the hybrid-labeling process can, in some cases, retain incorrect LLM extractions. The HRM prioritizes high-risk LLM outputs for manual correction, but lower-risk errors can persist in unflagged sources, and some high-value cases may potentially be missed. However, the combination of domain-informed filters, global and local outlier checks, and the Propose-Inspect-Review (PIR) post-processing pipeline adopted by the HUGO framework addresses many of these sources of error. For example, Kundu et al. \cite{kundu2025qualitative} report the tensile properties of cold spraying for repair, similar to many of the other studies here. However, whereas most repair-based studies remove the base part before testing, Kundu et al. performed tensile tests with the substrate as the spine of the sample, and a cold-sprayed coating on either side. The LLM incorrectly extracted these experiments as pure cold spray samples, and was only identified by a manual review of anomalous mechanical properties during downstream modeling.

\section{Conclusion}
This work introduces HUGO, a hybrid-labeling framework for extracting experimental values from literature and HUGO-CS, the largest machine-readable dataset of cold spray mechanical properties to date, containing 4,383 experiments over 1,124 articles with 144 features. HUGO combines LLM-based extraction with a Hierarchical Risk Mitigation (HRM) strategy tailored to common LLM failure modes, flagging high-risk outputs for targeted manual review. As a result, manual-labeling can be concentrated on extractions where the automated LLM-based extraction process is likely to fail to produce the correct label. This  improves extraction accuracy and coverage while still enabling high-throughput analysis. Additionally, to support downstream applications of HUGO-CS, such as advanced meta-reviews, process optimization, or model training, subsequent post-processing steps, such as categorical string mapping, continuous string processing, and unit normalization, were performed. Compared to prior work, to the best of our knowledge, HUGO-CS significantly expands the availability and accessibly of experimental cold spray results, and is released under an open-source license.  

\section{Data Availability}
All code to replicate this work, along with the complete HUGO-CS dataset, are released under a CC-BY license at \href{https://github.com/sprice134/HUGO}{https://github.com/sprice134/HUGO}.

\section{Acknowledgments}
The authors thank Aileen Cornwall-Brady, Yasmin Soares, Ezekiel Kasen, Camren McCormick, Wyatt Hazen, and Haley Talbot for their contributions and support with the hand-labeling process.

\section{Funding}
This research was developed with funding from the Defense Advanced Research Projects Agency (DARPA). The views, opinions, and/or findings expressed are those of the authors and should not be interpreted as representing the official views or policies of DARPA or the U.S. Government.

\section{Conflicts of Interest}
On behalf of all authors, the corresponding author states that there is no conflict of interest.

\bibliography{Ref.bib}
\bibliographystyle{spmpsci.bst}

\end{document}